\documentclass[sigconf]{acmart}

\copyrightyear{2025}
\acmYear{2025}
\setcopyright{cc}
\setcctype{by}
\acmConference[MM '25]{Proceedings of the 33rd ACM International Conference on Multimedia}{October 27--31, 2025}{Dublin, Ireland}
\acmBooktitle{Proceedings of the 33rd ACM International Conference on Multimedia (MM '25), October 27--31, 2025, Dublin, Ireland}
\acmISBN{979-8-4007-2035-2/2025/10}
\acmDOI{10.1145/3746027.3754707}

\AtBeginDocument{%
  }

\usepackage{multirow}
\usepackage{multicol}
\usepackage{algorithm}
\usepackage{algorithmic}

\usepackage{subcaption}

\acmSubmissionID{204}

\begin{document}

\title{Advancing Complex Wide-Area Scene Understanding with Hierarchical Coresets Selection}


\settopmatter{authorsperrow=4}
\author{Jingyao Wang}
\orcid{0000-0003-1782-8704}
\affiliation{%
  \institution{Institute of Software Chinese Academy of Sciences}
  \institution{University of the Chinese Academy of Sciences}
  \city{Beijing}
  \country{China}
}
\email{wangjingyao2023@iscas.ac.cn}

\author{Yiming Chen}
\orcid{0009-0005-0873-3078}
\affiliation{%
  \institution{Beijing University of Technology}
  \institution{Institute of Software Chinese Academy of Sciences}
  \city{Beijing}
  \country{China}
}
\email{chenyiming@emails.bjut.edu.cn}

\author{Lingyu Si}
\orcid{0000-0002-7735-6676}
\authornote{Corresponding author}
\affiliation{%
  \institution{Institute of Software Chinese Academy of Sciences}
  \institution{University of the Chinese Academy of Sciences}
  \city{Beijing}
  \country{China}
}
\email{lingyu@iscas.ac.cn}

\author{Changwen Zheng}
\orcid{0000-0002-2311-6757}
\affiliation{%
  \institution{Institute of Software Chinese Academy of Sciences}
  \institution{University of the Chinese Academy of Sciences}
  \city{Beijing}
  \country{China}
}
\email{changwen@iscas.ac.cn}


\renewcommand{\shortauthors}{Jingyao Wang, Yiming Chen, Lingyu Si, and Changwen Zheng}

\begin{abstract}
  Scene understanding is one of the core tasks in computer vision, aiming to extract semantic information from images to identify objects, scene categories, and their interrelationships. Although advancements in Vision-Language Models (VLMs) have driven progress in this field, existing VLMs still face challenges in adaptation to unseen complex wide-area scenes. To address the challenges, this paper proposes a Hierarchical Coresets Selection (HCS) mechanism to advance the adaptation of VLMs in complex wide-area scene understanding. It progressively refines the selected regions based on the proposed theoretically guaranteed importance function, which considers utility, representativeness, robustness, and synergy. Without requiring additional fine-tuning, HCS enables VLMs to achieve rapid understandings of unseen scenes at any scale using minimal interpretable regions while mitigating insufficient feature density. HCS is a plug-and-play method that is compatible with any VLM. Experiments demonstrate that HCS achieves superior performance and universality in various tasks. The code is available at \url{https://wangjingyao07.github.io/HCS.github.io/}.
\end{abstract}


\maketitle

\section{Introduction}
\label{sec:intro}

Scene understanding \cite{cordts2016cityscapes,li2009towards} is a fundamental task in computer vision, aiming to extract semantic information from images or videos to identify objects, scenes, and their interrelationships. Its significance is evident in various applications, e.g., autonomous driving \cite{jiang2023vad,neven2017fast}, intelligent surveillance \cite{zhang2012mining,ibrahim2016comprehensive}, remote sensing analysis \cite{NWPU-RESISC45,aarthi2017scene}, and medical diagnosis \cite{asgari2021deep,drew2013informatics}. Recent advances in vision-language models (VLMs) \cite{liao2024vlm2scene,liu2024vision,zhou2024embodied} further propelled it.

Despite their robust representational capabilities \cite{fu2024scene,cao2024maplm,wang2023awesome}, existing VLMs struggle with wide-area scenes. These scenes, spanning expansive environments such as deep-sea regions, remote sensing imagery, and areas with complex geographical features \cite{greene2009briefest,gu2019survey}, differ substantially from the urban or indoor settings typical of VLM pre-training \cite{naseer2018indoor,grant2017crowd}.
The challenges primarily arise from two interrelated factors. First, the extensive coverage inherent to wide-area scenes introduces an unprecedented diversity of object types (e.g., unknown species of sea snakes, ascidians, or sea urchins in oceans) that are rarely encountered in standard datasets. Second, although with various species, the data may exhibit a pronounced long-tail and sparse distribution in some areas. For instance, while sedimentary rocks might dominate the seabed, rarer features such as sponges appear only sporadically, leading to an overrepresentation of homogenized, high-frequency features.
One might argue that existing VLMs boast an abundance of parameters and deeper network architectures that facilitate the capture of intricate and global semantic information. However, these advantages do not inherently overcome the data challenges present in wide-area scenes. In fact, the models may overly focus on globally uniform, high-frequency features. Recent studies have proposed sample partitioning and multi-scale attention mechanisms \cite{shi2024vila,wang2024reasoning,zhang2024cls,wang2024image} to address these shortcomings; however, their effectiveness is heavily dependent on the quality of regional segmentation and scale priors, and they often neglect the relationships between different regions. Furthermore, the substantial computational resources required to process the massive volumes of visual data further restrict the practical deployment of such models.
Therefore, while pre-training endows VLMs with impressive capabilities, the unique challenges of wide-area scenes pose significant obstacles to their effective adaptation.

To tackle this, we aim to propose a method that can adaptively refine the relationships between scene regions and model decisions in a coarse-to-fine manner for complex wide-area scene understanding. This allows the VLMs to quickly grasp scenes of any scale or complexity using only a few interrelated and interpretable regions.
We begin by investigating the interpretable regional characteristics that support effective selection in wide-area scenes. Inspired by data compression theory \cite{storer1987data,sayood2017introduction}, we propose that the essence of feature selection is to find a small, weighted subset of the original dataset, ensuring that learning tasks on this subset yields an approximately optimal solution. Accordingly, accurate scene understanding requires that the model establish appropriate evaluation criteria tailored to specific tasks, and filter out regions that critically influence decision-making to achieve the key subset, i.e., interpretable regions. 
To achieve this, we incorporate coreset theory \cite{har2004coresets}, which enables the construction of a compact subset of input data, i.e., coreset, that serves as a proxy for the whole dataset while maintaining minimal deviation in accuracy. By reducing the data volume, coresets enhance scalability and efficiency. A common approach for constructing coresets involves sampling data points with probabilities proportional to their so-called sensitivity scores, which quantify the worst-case impact of a point on a given attribute of interest \cite{phillips2017coresets}. Take a classification task as an example, this attribute is often a smooth convex approximation of the misclassification rate \cite{huang2024optimal}. However, the ultimate objective in wide-area scene understanding extends beyond only optimizing the loss function on observed training data to understanding classifier performance on unseen data.
This leads to two limitations of existing coreset selection strategies: (i) their effect-only selection, bounded by fixed partitions, often produces regions that are too large or sparse, impairing optimization; and (ii) they typically ignore inter-region relationships, reducing prediction accuracy.

To address the aforementioned issue, we propose a plug-and-play Hierarchical Coresets Selection mechanism (HCS). 
It leverages a theoretically guaranteed importance function to weight regions and employs a layer‐by‐layer refinement strategy for precise coreset selection.
Specifically, initially, HCS partitions the samples coarsely using the feature significance map of baseline VLMs. It then leverages the importance function that evaluates the interpretability of different regions along four key dimensions: utility, representativeness, robustness, and synergy. This metric not only accounts for inter-regional relationships but also accurately attributes error sources in misclassified samples—distinguishing it from traditional performance-only methods.
To alleviate the insufficient density issue in the attributed regions, we further refine the selected regions, i.e., partitioning the regions at a smaller scale based on the reweighting significance map. By iterating the above steps, HCS enables the model to rely on only a few interpretable regions to achieve a rapid understanding of scene images of any scale, without the need for additional training.
HCS can be embedded into any VLM. Extensive experiments across multiple tasks and a wide range of VLM baselines demonstrate the superiority and universality of HCS.
In summary, the main contributions can be summarized as:
\begin{itemize}
    \item We explore a more challenging problem of complex wide-area scene understanding, and reformulate it as a coreset feature selection problem, aiming for accurate and stable understanding with fewer interpretable regions. 
    \item We propose a hierarchical coreset selection mechanism (HCS) for precise wide-area scene understanding. It employs a theoretically validated importance function considering utility, representativeness, robustness, and synergy to assign weights and implements a layer‐by‐layer refinement strategy for coreset selection. 
    HCS is a plug-and-play method that allows any VLM to achieve training-free scene understanding with only a few interpretable regions. 
    \item Extensive experiments on various datasets demonstrate the effectiveness of HCS on VLMs for scene understanding.
\end{itemize}

\section{Related Work}
\label{sec:related_work}

\subsection{Scene Understanding}
Scene understanding \cite{cordts2016cityscapes,li2009towards,neven2017fast,aarthi2017scene,wang2022so} aims to analyze objects, semantics, and spatial relationships in complex environments, serving as a fundamental component of intelligent perception and decision-making. Traditional approaches \cite{qi2016dynamic,hong2018d3,huang2018long,peng2019part,wang2023amsa,qiang2024universality} rely heavily on handcrafted feature descriptors, but recent breakthroughs in deep learning have significantly accelerated progress in this field, particularly VLMs \cite{fu2024scene,cao2024maplm,qi2025gpt4scene}.
CLIP2Scene \cite{chen2023clip2scene} and VLM2Scene \cite{liao2024vlm2scene} leverage CLIP to improve scene understanding for autonomous driving. Cao et al. \cite{cao2024maplm} proposed MAPLM, a multimodal instruction-tuning baseline that integrates CLIP with LLaMA-2/Vicuna for scene understanding; Wang et al. \cite{wang2024tsp} introduced TSP-Transformer, incorporating task-specific prompts onto VLMs for scene understanding. However, these approaches remain limited to structured, small-scale environments such as rooms and intersections. They also heavily depend on high-quality data for inference and adaptation, making it difficult to generalize effectively to wide-area unstructured scenarios. To address this, in this paper, we propose a hierarchical coreset selection (HCS) method that refines the model’s understanding across four key dimensions: utility, representativeness, robustness, and synergy. By progressively identifying a minimal set of interpretable regions and their interactions, it enables VLMs to activate their capabilities in new tasks, supporting accurate and efficient wide-area scene understanding.

\subsection{Coreset Theory}
Coreset theory \cite{storer1987data,sayood2017introduction} is a data compression theory designed to efficiently reduce large-scale datasets into smaller subsets while preserving key statistical properties. By retaining essential information, it helps lower computational complexity, which is particularly effective for high-dimensional data. 
Recent studies \cite{tukan2020coresets,sener2017active,wang2024towards} have integrated Coreset with deep learning, using it to evaluate data quality by quantifying how selected samples impact model performance. For example, Sener et al. \cite{sener2017active} used the k-Center algorithm to select representative subsets while maintaining model performance on the remaining data. Similarly, Vepa et al. \cite{vepa2025integrating} applied layer-wise active learning strategies to reduce annotation costs and improve medical image detection efficiency using Coreset selection.
In this paper, we leverage Coreset theory to optimize VLM performance in wide-area scene understanding tasks. Unlike previous Coreset selection \cite{har2004coresets,huang2024optimal}, which primarily focuses on data utility, we consider four key factors, i.e., utility, representativeness, robustness, and synergy, and propose a theoretically grounded selection framework that ensures interpretable region selection for any scene scale.

\begin{figure*}
\begin{subfigure}{0.7\linewidth}
        \centering
        \includegraphics[width=\textwidth]{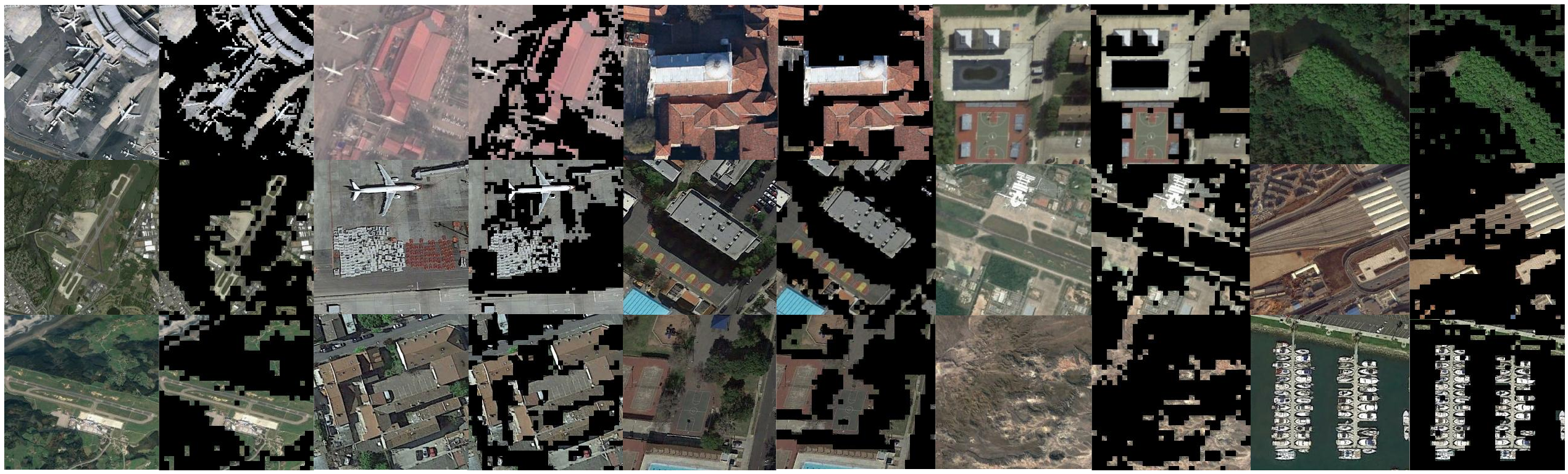}
        \caption{Visualization of original samples and region selection}
        \vspace{-0.1in}
    \end{subfigure}
    \hfill
    \begin{subfigure}{0.275\linewidth}
        \centering
        \includegraphics[width=\textwidth]{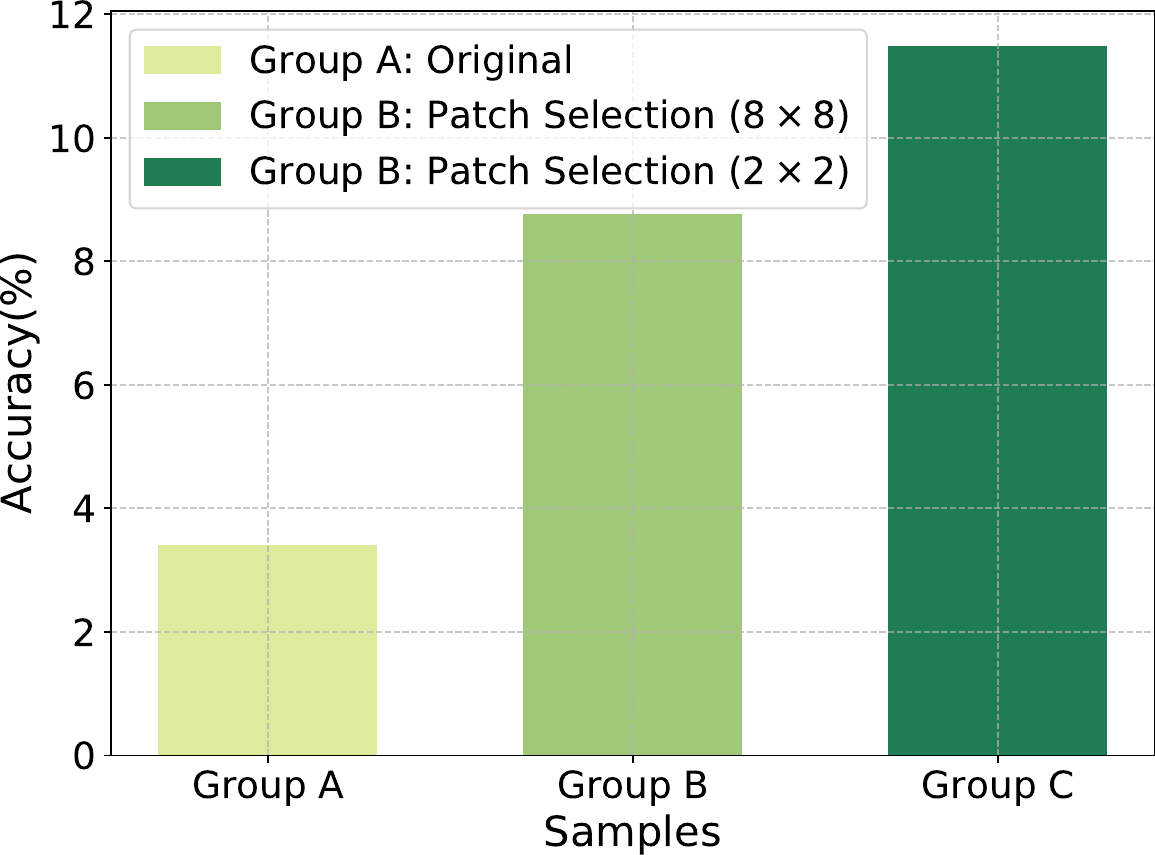}
        \caption{Performance comparison} 
        \vspace{-0.1in}
    \end{subfigure}
    \caption{Motivating results on NWPU-RESISC45. (a) shows the visualization results of regions selected via coreset performance (following Eq.\ref{eq:coreset_rule}). (b) shows the performance of the pre-trained model on the original samples and after region selection.}
    \label{fig:motivation}
\end{figure*}

\subsection{Vision-Language Models (VLMs)}
VLMs undergo pre-training to align visual and textual representations. Recent research on VLMs \cite{li2024clipsam,xie2020pointcontrast,wang2023amsa,wang2024causal} has achieved great generalization capabilities on various applications, including scene understanding \cite{chen2023clip2scene,liao2024vlm2scene} that this paper focuses on. Unfortunately, VLMs face significant challenges in adapting to complex, wide-area scenes, e.g., deep sea, despite their strong representational capabilities on common scenes, e.g., in-door rooms \cite{naseer2018indoor,grant2017crowd}. The wide-area scenes present greater diversity in object types, broader spatial ranges, and more pronounced scale variations, making accurate scene understanding difficult. Some methods \cite{aarthi2017scene,chen2025advancinggeneralmultimodalcapability} proposed to expand the receptive field to enhance global context comprehension, but this often comes at the cost of losing critical local details. While multi-scale attention mechanisms \cite{shi2024vila,wang2024reasoning} can help extract richer features, the massive scale of visual information in wide-area scenes leads to significantly higher computational costs. To ensure fast and precise wide-area scene understanding, this work proposes to select and refine the interpretability region layer by layer adaptively to activate the representational capability of VLMs.

\section{Problem Analysis and Motivation}
\label{sec:problem_analysis}
In this section, we begin by introducing the problem settings and notations of wide-area scene understanding, specifically, what defines the coresets under this setting (\textbf{Subsection \ref{sec:problem_settings}}). Next, we provide empirical evidence about the importance of our coreset concepts and the limitations of existing VLMs (\textbf{Subsection \ref{sec:empirical_evidence}}).

\subsection{Problem Settings}
\label{sec:problem_settings}
Let \(X \in \mathbb{R}^{H \times W \times C}\) denote a wide-area scene image (with height \(H\), width \(W\), and \(C\) channels) and \(Y\) the corresponding ground truth (e.g., semantic labels or detection boxes). The goal is to learn a mapping $f_\theta: X \rightarrow Y$ parameterized by \(\theta\), by minimizing the loss $\min_{\theta} \mathcal{L}(f_\theta(X), Y)$.
Due to the vast and heterogeneous nature of wide-area scenes, the image \(X\) can be decomposed into \(\mathcal{X}\), the set of minimal units (\(x\in \mathcal{X}\), e.g., small patches) extracted from wide‐area scene images, and that \(P\) is a probability measure over \(\mathcal{X}\).
Let \(\Theta\) denote the parameter space of candidate models, and \(\ell: \mathcal{X} \times \Theta \to [0,\infty)\) be a loss function defined on a single unit \(x\). The full loss evaluated at a model parameter \(\theta \in \Theta\) is given by
\begin{equation}\label{eq:coreset_l_problem}
    \mathcal{L}(\theta) = \int_{x \in \mathcal{X}} \ell(x,\theta) \, \mathrm{d}P(x).
\end{equation}
A finite set \(X \subseteq \mathcal{X}\), together with a weight function \(\nu: X \to \mathbb{R}_{\ge 0}\), is termed an \(\epsilon\)-coreset for \((\mathcal{X},P,\Theta,\ell)\) (with \(\epsilon \in (0,1)\)) if it approximates the full loss for every \(\theta \in \Theta\) as follows:
\begin{equation}\label{eq:coreset_rule}
\left| \mathcal{L}(\theta) - \sum_{x \in X} \nu(x) \, \ell(x,\theta) \right| \le \epsilon\, \mathcal{L}(\theta).
\end{equation}
where each \(x \in X\) represents a minimal image unit from the wide-area scene, ensuring that the coreset effectively approximates the overall loss while capturing fine-grained local details.

\subsection{Empirical Evidence}
\label{sec:empirical_evidence}
To evaluate the performance of VLMs in wide-area scene understanding, particularly their ability to capture fine-grained information, we design a set of experiments in this subsection. Our goal is to determine whether pre-trained VLMs, without any fine-tuning, can fully comprehend the critical local details in wide-area scenes and whether coreset selection can enhance their performance.

Specifically, we select the RESISC45 dataset \cite{NWPU-RESISC45} as our benchmark for scene classification. This is a remote sensing dataset comprising 45 scene categories, with image samples exhibiting spatial resolutions ranging from 20 centimeters to over 30 meters per pixel. We divide the dataset into three resolution groups—high, medium, and low—based on ground sample distance (GSD) using the one-third quantiles. We use a pre-trained CLIP model as our baseline following the settings in \cite{CLIP}. We then conduct two experimental setups to assess model performance. In the first setup, the entire high-resolution image is directly fed into the CLIP model, and we record its zero-shot performance on scene classification. In the second setup, we employ a region selection strategy: each image is partitioned into fixed-size patches, each patch is scored for its impact on the final loss (Eq.\ref{eq:score_ut}), and only the most contributive regions are selected for feature fusion before being passed to the CLIP model for prediction. To account for the influence of patch size on performance, we experiment with two configurations: \(2 \times 2\) and \(8 \times 8\) patches. Both approaches rely solely on the pre-trained model without any fine-tuning, which allows us to directly compare whole-image input and region selection strategies.

The experimental results, as shown in \textbf{Figure \ref{fig:motivation}}, reveal that: (i) incorporating the region selection strategy significantly improves classification accuracy compared to directly using the entire image as input; (ii) the optimal proportion of candidate regions varies across images with different resolutions, where the selected size impacts performance. These findings suggest that, in wide-area scene understanding tasks, despite the large capacity and strong representational abilities of current VLMs, their adaptability remains limited. Meanwhile, coreset selection effectively enhances model performance by making them focus on critical local details. Moreover, as the optimal selection strategy varies across data, these results motivate us to dynamically adjust region sizes based on data characteristics to support more accurate scene understanding.

\begin{figure*}
    \centering
    \includegraphics[width=0.93\textwidth]{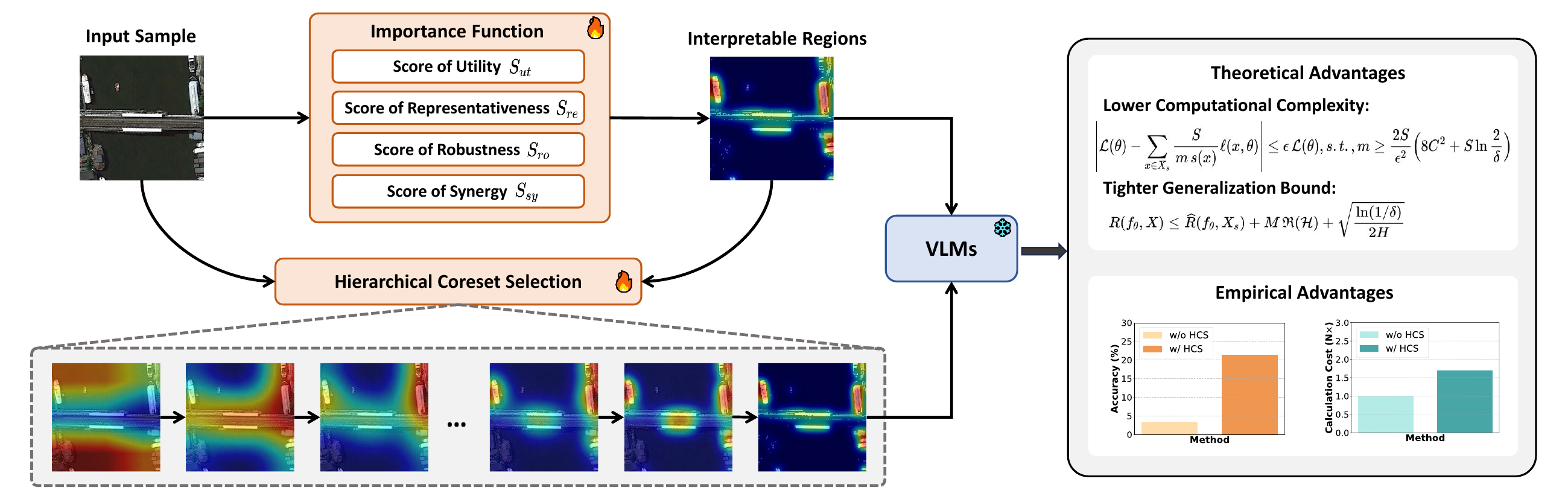}
    \vspace{-0.12in}
    \caption{The framework of the proposed HCS. Its integration allows VLMs, to be tested without fine-tuning (frozen), instead training this lightweight network (HCS) for coreset selection to enhance model performance.}
    \vspace{-0.1in}
    \label{fig:hcs}
\end{figure*}

\section{Methodology}
\label{sec:method}
Based on the above insights, in this paper, we aim to develop a dynamic coreset selection method so that a small subset of interpretable regions can serve as a proxy for the entire dataset. This approach ensures that the predictive performance of VLMs even surpasses that obtained with the full set, thereby effectively activating the capability of VLMs for wide-area image understanding.
Generally, existing coreset theory mainly relies on a sensitivity-based sampling strategy, where data points are sampled in proportion to their influence on the loss function. However, for wide-area scene understanding, this strategy faces two major limitations: (i) the construction of the coreset depends on prior partitioning, which may lead to selected regions being insufficiently fine-grained (i.e., the regions are too large or not dense enough), thus interfering with the fine-grained optimization; (ii) effect-based selection methods struggle to accurately attribute the causes of prediction errors to specific misclassified samples or regions while neglecting the interdependence among different regions, affecting performance.

To address these issues, in this section, we propose a plug-and-play Hierarchical Coreset Selection (HCS) mechanism. HCS employs a theoretically grounded importance function and a layer-wise refinement strategy for precise coreset selection. Specifically, considering the characteristics of wide-area scene images, HCS redefines the evaluation metrics for coresets by establishing an importance function from four perspectives—utility, representativeness, robustness, and synergy—to support accurate coreset selection (\textbf{Subsection \ref{sec:method_function}}). The theoretical analysis in \textbf{Section \ref{sec:theoretical}} demonstrates its superiority. Moreover, HCS partitions regions through iterative refinement, segmenting them from coarse to fine scales based on the obtained variable-scale feature importance maps at each iteration (\textbf{Subsection \ref{sec:method_selection}}). With only a few steps, HCS enables the VLMs to rapidly comprehend scene images of any scale using only a small number of interpretable regions, without requiring additional training. The framework of HCS is shown in \textbf{Figure \ref{fig:hcs}}.

\subsection{Importance Function Design}
\label{sec:method_function}

In wide-area scene understanding, it is essential that the selected coreset not only independently provides critical information but also works synergistically with other regions to enhance overall model performance. To achieve this, we incorporate four key metrics—utility, representativeness, robustness, and synergy—which comprehensively assess the importance of each region from multiple dimensions. They constitutes the importance function for coreset selection. Specifically, we define the coreset as $X_{s}$, a subset of the full image $X$. Then, the four metrics are defined as follows:

\textbf{Score of Utility} 
 measures the impact of the coreset $X_s$ on the overall loss and the spatial compactness of $X_s$, i.e., only a few regions are included as the proxy of the whole sample. It is defined as:
 \begin{equation}\label{eq:score_ut}
     S_{ut}(X_s)= \exp\left(- \frac{\|\mathcal{L}(X) - \mathcal{L}(X_s)\|}{\mathcal{L}(X)}\right) + \lambda\left(1 - \frac{A(X_s)}{A(X)}\right),
 \end{equation}
 where \(\mathcal{L}(\cdot)\) denotes the loss, \(A(\cot)\) is the area operator, and $\lambda$ is the weight balancing losses ($\lambda=0.7$ in this paper through parameter sensitivity experiments). A higher \(S_{ut}(X_s)\) indicates that the coreset \(X_s\) is highly informative and occupies only a minimal portion of $X$.

\textbf{Score of Representativeness} evaluates how well the feature distribution of the coreset $X_s$, denoted as \(p(X_s)\), matches the global feature distribution \(p(X)\). It is defined as:
\begin{equation}\label{eq:score_re}
    S_{re}(X_s)=\exp(- \frac{D_{\mathrm{KL}}\bigl(p(X_s)\,\|\,p(X)\bigr)}{D_{\max}})
\end{equation}
where \(D_{\mathrm{KL}}(\cdot\|\cdot)\) denotes the Kullback–Leibler divergence, and \(D_{\max}\) is a normalization constant representing the maximum expected divergence. When the coreset distribution closely approximates the global distribution, resulting in a near-zero \(D_{\mathrm{KL}}\) and an \(S_{re}(X_s)\) approaching 1. Conversely, if the coreset distribution deviates significantly from the global distribution (for example, by neglecting rare but important information due to long-tail effects), \(D_{\mathrm{KL}}\) increases, causing the exponential function to decay rapidly and yielding a low representativeness score. This score aims to mitigate bias induced by long-tail distributions within the wide-area scene samples.

\textbf{Score of Robustness} quantifies the stability of the features of the selected coreset $X_s$ under small perturbations. For example, remote-sensing images may face color fluctuations or speckle noise issues during the acquisition process. This score is defined as:
\begin{equation}\label{eq:score_ro}
    S_{ro}(X_s)=\exp(- \frac{\|f_\theta(X_s) - f_\theta(X_s+\delta)\|}{\delta_{\max}}),
\end{equation}
where \(f_\theta(X_s)\) is the feature output for region \(X_s\) by the fixed model \(f_\theta\), \(\delta\) represents a small perturbation applied to \(X_s\), and \(\delta_{\max}\) is a reference maximum change used for normalization. When the features remain nearly unchanged under perturbation, the norm \(\|f_\theta(X_s) - f_\theta(X_s+\delta)\|\) is small, causing the exponential term to be close to 1, which indicates high robustness. It ensures the reliability of selected coreset $X_s$ even in noisy or variable conditions.

\textbf{Score of Synergy} measures the complementary effect of regions within $X_s$ when combined with each other. Given a region \(r' \in X\setminus X_s\)), the score can be expressed as:
\begin{equation}\label{eq:score_sy}
    S_{sy}(X_s)=\exp\left(-\left|\frac{\mathcal{L}(X) - \mathcal{L}\bigl(X \setminus (X_s \cup \{r'\})\bigr)}{\mathcal{L}(X) - \mathcal{L}\bigl(X \setminus \{r'\}\bigr)}\right|\right)
\end{equation}
where \(\mathcal{L}({X-X_s\cup\{r'\}})\) is the loss when remove the entire coreset and \(r\), and \(\mathcal{L}({X-\{r'\}})\) is the loss when only region \(r'\) is removed. 
If the exponential term approaches 1, then removing \(r'\) in combination with \(X_s\) yields nearly the same loss reduction as removing \(r'\) alone. In other words, the current coreset \(X_s\) already captures the information of \(r'\), indicating that \(X_s\) is nearly optimal. 
It aligns with the goal of selecting a coreset that maintains the overall performance while being as compact and non-redundant as possible. Meanwhile, it considers the relationships across different regions that previous methods ignore for wide-area scene understanding.

Based on the above four metrics, we design the importance function for coreset selection. The overall score for each region is a combination of the four components, which can be expressed as: 
\begin{equation}\label{eq:score}
    S(X_s)=S_{ut}(X_s)+\lambda_{re} S_{re}(X_s)+ \lambda_{ro}S_{ro}(X_s)+\lambda_{sy} S_{sy}(X_s)
\end{equation}
where $\lambda_{re}$, $\lambda_{ro}$, and $\lambda_{sy}$ are the weights of the terms corresponding to representativeness $S_{re}(\cdot)$, robustness $S_{ro}(\cdot)$, and synergy $S_{sy}(\cdot)$. All weighting coefficients are chosen through parametric search (as shown in \textbf{Subsection \ref{sec:ex_4_ablation}}, for simplicity, they can be set to 1 by default with a performance loss of no more than 2\%). This function enables us to rank candidate regions and select those with the highest scores as the coreset. By doing so, the selected regions collectively capture the most critical, diverse, stable, and complementary information of the full data, thereby achieving an efficient and effective proxy for understanding the wide-area scene.

\subsection{Hierarchical Coresets Selection}
\label{sec:method_selection}

To obtain accurate coresets from wide-area images, HCS performs hierarchical selection with the aforementioned importance function. Unlike traditional methods \cite{redmon2018yolov3,wang2023amsa,zhan2025coreset} that rely on fixed patch partitioning strategies, we employ a layer-wise refinement strategy, first selecting high-scoring regions at a coarse granularity and then conducting refined searches within these candidate regions.

Specifically, we first uniformly divide the input image \( X \in \mathbb{R}^{w \times h \times c} \) into \( N_p \times N_p \) large-scale patches (set as \(4 \times 4\) in this work), with each patch considered as a candidate region, denoted as \( \mathcal{R} = \{r_{ij}\}, i,j=1,\dots,N \).
Next, we compute a score \( S(r_{ij}) \) for each region \( r_{ij} \) based on a predefined importance function \( S(\cdot) \) (see Eq.\ref{eq:score}). This function comprehensively integrates factors such as utility, representativeness, robustness, and synergy effects. We then rank all candidate regions by their scores and select the top \( K \) regions to form an initial candidate set \( \mathcal{R}_s \). During this selection, we simultaneously verify that each chosen region satisfies the core set definition, ensuring the resulting core set closely approximates the loss of the entire image, i.e., ensuring $|\mathcal{L}(X) - \mathcal{L}(X_s)| \le \epsilon \, \mathcal{L}(X)$ ($\epsilon=0.98$ in this work).
After identifying the preliminary candidate set, we further subdivide each selected region \( r \in \mathcal{R}_s \) into patches and repeat the evaluation and selection process described above. Notably, since we have introduced a synergistic scoring component $S_{sy}$, each chosen region's combined performance with adjacent patches is explicitly assessed. This strategy facilitates boundary refinement and ensures that the final selected regions collectively meet the core set requirements.
Finally, we aggregate the finely screened candidate regions obtained within each coarsely selected patch to construct the final core set \( X_s = \bigcup_{r \in \mathcal{R}_{s}} r \). Based on this, HCS achieves an efficient selection process that progressively refines from the full image to local regions, from coarse to fine, obtaining a precise coreset for accurate wide-area scene understanding.

\section{Theoretical Analysis}
\label{sec:theoretical}
In this section, we conduct theoretical analyses to demonstrate that HCS can significantly reduce the computational complexity while achieving better generalization.
Specifically, we begin by demonstrating that the coreset selected by our HCS is bounded and effective, achieving complexity reduction (\textbf{Theorem \ref{theo:1}}). Next, we prove that the selected coreset not only guarantees an approximation of the full loss, but also yields a tighter generalization upper bound (\textbf{Theorem \ref{theo:2}}). The proofs are shown in \textbf{Appendix A}.

We begin by showing that the coreset selected by our HCS is both bounded and helps reduce overall complexity.
\begin{theorem}\label{theo:1}
    Let \(X \in \mathbb{R}^{w\times h\times c}\) be a wide-area scene image and \(\mathcal{X}\) be the set of minimal units extracted from \(X\). Let \(P\) be a probability measure on \(\mathcal{X}\) and define the loss of a fixed model \(f_\theta\) on \(X\) as Eq.\ref{eq:coreset_l_problem}. Assume there exists an upper importance function \(s:\mathcal{X}\to (1,\infty)\) for \(\ell(\cdot,\theta)\) with total score $S = \int_{\mathcal{X}} s(x)\,dP(x)$ for $X_s$. Through HCS, if we select \(X_s\subset \mathcal{X}\) that satisfy $m\ge g(\epsilon,S,\delta)=\frac{2S}{\epsilon^2}\Bigl(8C^2+S\ln\frac{2}{\delta}\Bigr)$ and assign weights $\nu(x)=\frac{S}{m\,s(x)}$, then with probability at least \(1-\delta\) it holds for all \(\theta\) achieve $\left|\mathcal{L}(\theta)-\sum_{x\in X_s} \nu(x)\ell(x,\theta)\right|\le \epsilon\,\mathcal{L}(\theta)$.
\end{theorem}
This theorem shows that \(X_s\) is bounded in the sense that its weighted loss estimate approximates the full loss \(\mathcal{L}(\theta)\) within an \(\epsilon\) fraction. The guarantee implies that even though \(X_s\) is a small subset of \(\mathcal{X}\), it retains sufficient information to closely approximate the performance of the model on the full image. Meanwhile, it also indicates a reduction in overall complexity, i.e., the size \(m\) is determined by the function \(g(\epsilon,S,\delta)\), which depends on the tolerance \(\epsilon\), the total sensitivity \(S\), and the confidence parameter \(\delta\).

Next, we turn to analyze the performance of $f_\theta$ with coresets.
\begin{theorem}\label{theo:2}
    Let the hypothesis class of the model be $\mathcal{H}\subseteq\{h:\mathbb{R}^d\to \mathcal{Y}\}$ with pseudo-dimension $H = \mathrm{Pdim}(\{\ell_h : h\in\mathcal{H}\})$, and let \(\mathfrak{R}(\mathcal{H})\) denote the Rademacher complexity of \(\mathcal{H}\). Suppose the coreset \(X_s\) selected via HCS satisfies $\left|\mathcal{L}(X)-\mathcal{L}(X_s)\right|\le \epsilon\,\mathcal{L}(X)$. Then, for a fixed model \(f_\theta\) and samples drawn from the empirical distribution \(\mathcal{D}_{tr}\), with probability at least \(1-\delta\) the following bound holds:
    \begin{equation}
        R(f_\theta,X) \le \widehat{R}(f_\theta, X_s) + M\,\mathfrak{R}(\mathcal{H}) + \sqrt{\frac{\ln(1/\delta)}{2H}},
    \end{equation}
    where \(R(f_\theta,X)\) denotes the risk on the full image \(X\), \(\widehat{R}(f_\theta, X_s)\) is the empirical risk computed on the coreset \(X_s\), \(M\) is a finite constant.
\end{theorem}
This theorem demonstrates that the coreset selected via HCS not only approximates the full-image loss within the tolerance \(\epsilon\) but also, when combined with standard generalization error terms based on Rademacher complexity and pseudo-dimension, yields a tighter overall upper bound. Together, these theorems show that the HCS mechanism produces a coreset \(X_s\) that is both bounded—guaranteeing that the loss approximation error is within \(\epsilon\)—and enables a tighter generalization bound when evaluated with respect to the true risk. By leveraging the proposed importance function, HCS achieves significant complexity reduction and provides robust theoretical guarantees for wide-area scene understanding.

\begin{table*}
    \centering
    \caption{Performance comparison (Accuracy \%) of scene image classification on NWPU-RESISC45, AID, and RSI-CB. 
    Unless otherwise specified, we directly use the pre-trained models without fine-tuning and evaluate their transfer performance. 
    The brackets ``()'' indicate the effect changes of vanilla baselines after introducing HCS. 
    More details are shown in Appendix \ref{sec_app:experiment}.}
    \vspace{-0.1in}
    \label{tab:classification_results}
    \resizebox{1\linewidth}{!}{
    \begin{tabular}{lcccccc}
        \toprule
        \multirow{2}{*}{Model} & \multicolumn{2}{c}{NWPU-RESISC45} & \multicolumn{2}{c}{AID} & \multicolumn{2}{c}{RSI-CB} \\
        \cmidrule(lr){2-3} \cmidrule(lr){4-5} \cmidrule(lr){6-7}
         & Top-1 ACC & Top-5 ACC & Top-1 ACC & Top-5 ACC & Top-1 ACC & Top-5 ACC \\
        \midrule
        ViT-B/32 & 1.84 & 4.61 & 0.97 & 4.63 & 6.15 & 15.33 \\
        ViT-B/32+HCS & 17.15 (+15.31) & 28.56 (+23.95) & 12.17 (+11.20) & 20.61 (+15.98) & 19.51 (+13.36) & 27.64 (+12.31) \\
        \midrule
        ViT-L/14 & 2.23 & 6.85 & 1.37 & 5.53 & 8.89 & 17.26 \\
        ViT-L/14+HCS & 19.23 (+17.00) & 29.00 (+22.15) & 15.12 (+13.75) & 21.59 (+16.06) & 21.32 (+12.43) & 31.88 (+14.62) \\
        \midrule
        CLIP & 3.41 & 12.19 & 2.16 & 7.97 & 12.32 & 29.17 \\
        CLIP+HCS & 21.36 (+17.95) & 36.78 (+24.59) & 18.22 (+16.06) & 27.91 (+19.94) & 31.39 (+19.07) & 47.05 (+17.88) \\
        \midrule
        ContextCLIP & 1.48 & 11.05 & 1.01 & 6.25 & 11.32 & 25.88 \\
        ContextCLIP+HCS & 18.56 (+17.08) & 33.12 (+22.07) & 16.01 (+15.00) & 24.37 (+18.12) & 29.46 (+18.14) & 44.95 (+19.07) \\
        \midrule
        LLaVA-hf/llava-v1.6-mistral-7b-hf & 41.20 & 52.15 & 35.13 & 49.36 & 52.17 & 59.30 \\
        LLaVA-hf/llava-v1.6-mistral-7b-hf+HCS & 44.15 (+2.95) & 54.48 (+2.33) & 40.05 (+4.92) & 54.13 (+4.77) & 55.66 (+3.49) & 65.59 (+6.29) \\
        \midrule
        LLaVA-hf/llama3-llava-next-8b-hf & 53.12 & 64.84 & 39.65 & 51.06 & 58.33 & 64.89 \\
        LLaVA-hf/llama3-llava-next-8b-hf+HCS & 58.42 (+5.30) & 69.15 (+4.31) & 42.05 (+2.40) & 54.89 (+3.83) & 60.50 (+2.17) & 68.36 (+3.47) \\
        \midrule
        Qwen/Qwen2-VL-7B-Instruct & 61.15 & 75.56 & 46.26 & 62.03 & 68.63 & 76.69 \\
        Qwen/Qwen2-VL-7B-Instruct+HCS & 66.12 (+4.97) & 80.38 (+4.82) & 51.04 (+4.78) & 68.11 (+6.08) & 76.19 (+7.56) & 80.97 (+4.28) \\
        \bottomrule
    \end{tabular}}
\end{table*}

\begin{table}
    \centering
    \caption{Performance comparison (dice score \%) of scene semantic segmentation on TikTok dances, TrashCan, and GTEA. Each dataset is split into annotated base and unannotated target classes in a 1:1 ratio, and models are fine-tuned on the base class from the pre-trained weights. The brackets ``()'' indicate the effect changes after introducing HCS.}
    \label{tab:segmentation_results}
    \vspace{-0.1in}
    \resizebox{\linewidth}{!}{
    \begin{tabular}{lccc}
        \toprule
        Model & TikTok dances & TrashCan & GTEA \\
        \midrule
        Vanilla SAM     & 38.13 & 21.15 & 28.12 \\
        Vanilla SAM+HCS & 42.69 (+4.56) & 25.88 (+4.73) & 32.01 (+3.89) \\
        \midrule
        Med-SA        & 47.05 & 29.81 & 36.69 \\
        Med-SA+HCS    & 50.01 (+2.96) & 33.74 (+3.93) & 39.26 (+2.57) \\
        \midrule
        SAMed         & 49.11 & 30.26 & 35.90 \\
        SAMed+HCS     & 53.24 (+4.12) & 35.91 (+5.65) & 40.02 (+4.12) \\
        \midrule
        BLO-SAM       & 62.82 & 43.55 & 49.07 \\
        BLO-SAM+HCS   & 65.29 (+2.47) & 49.04 (+5.49) & 53.71 (+4.64) \\
        \midrule
        OVSAM         & 60.89 & 44.65 & 50.38 \\
        OVSAM+HCS     & 64.03 (+3.14) & 47.91 (+3.26) & 52.29 (+1.91) \\
        \midrule
        CLIPSAM       & 59.81 & 41.22 & 48.56 \\
        CLIPSAM+HCS   & 63.01 (+3.20) & 48.93 (+6.71) & 50.03 (+1.47) \\
        \bottomrule
    \end{tabular}}
\end{table}

\begin{table}[t]
\centering
\caption{Impact of different components in importance function on NWPU-RESISC45. ``$\bullet$'' and ``$\circ$'' indicate the availability and absence of the corresponding terms. The brackets ``()'' indicate the effect changes after introducing HCS.}
\label{tab:abla_1}
\vspace{-0.1in}
\setlength\tabcolsep{1pt}
\begin{center}
\begin{tabular}{cccc|cc}
\toprule
\multicolumn{4}{c|}{Component} & \multicolumn{2}{c}{Performance} \\ 
\midrule
$S_{ut}$ & $S_{re}$ & $S_{ro}$ & $S_{sy}$
& Top-1 Accuracy (\%) & Top-5 Accuracy (\%) \\ 
\midrule
$\bullet$ & $\circ$ & $\circ$ & $\circ$ & 15.39 (+11.98) & 23.51 (+11.32)\\
$\bullet$ & $\bullet$ & $\circ$ & $\circ$ & 17.26 (+13.85) & 28.89 (+16.70) \\
$\bullet$ & $\circ$ & $\bullet$ & $\circ$ & 16.97 (+13.56) & 26.45 (+14.26) \\
$\bullet$ & $\circ$ & $\circ$ & $\bullet$ & 18.05 (+14.64) & 28.65 (+16.46) \\
$\bullet$ & $\bullet$ & $\bullet$ & $\circ$ & 18.21 (+14.80) & 31.44 (+19.25) \\
$\bullet$ & $\bullet$ & $\circ$ & $\bullet$ & 20.01 (+16.60) & 33.47 (+21.28) \\
$\bullet$ & $\circ$ & $\bullet$ & $\bullet$ & 18.93 (+15.52) & 32.48 (+20.29) \\
$\bullet$ & $\bullet$ & $\bullet$ & $\bullet$ & 21.36 (+17.95) & 36.78 (+24.59) \\ 
\bottomrule
\end{tabular}
\end{center}
\end{table}

\section{Experiments}
\label{sec:experiment}
In this section, we conduct extensive experiments on various
benchmark datasets to evaluate the effectiveness of HCS. More details and experiments are provided in \textbf{Appendices \ref{sec_app:dataset}-\ref{sec_app:experiment}}.

\subsection{Experimental Settings}
\label{sec:ex_1_settings}
In this subsection, we introduce the datasets, baselines, implementation details, and evaluation metrics of our experiments in turn.

\textbf{Benchmark Datasets.}
We select two scene understanding tasks, namely scene image classification and semantic segmentation, covering six datasets. For scene image classification, we evaluate three benchmark datasets:
(i) NWPU-RESISC45 \cite{NWPU-RESISC45}: contains 31,500 images covering 45 scene categories, with various resolutions for multi-scale remote sensing image classification.
(ii) AID \cite{AID}: consists of over 10,000 aerial images covering 30 categories, collected from different regions with high diversity, and is used to assess the generalization effect of HCS on VLMs.
(iii) RSI-CB \cite{RSI-CB}: includes approximately 36,000 scene image patches across 45 categories, which can simulate wide-area scene classification by stitching together similar types.
For semantic segmentation, we select three benchmark datasets:
(i) TikTok dances \cite{roman2023humantiktok}: comprises 2,615 images of dancing individuals extracted from TikTok videos, with full-body segmentation. 
(ii) TrashCan \cite{hong2020trashcan}: contains 1,484 real underwater images labeled with six types of waste. Due to the challenges posed by underwater lighting, this dataset is used to evaluate the improvement in model robustness with HCS.
(iii) GTEA \cite{fathi2011learning}: comprises 7 sets of first-person view daily activity scene videos (3,500 frames), featuring occlusions and complex human movements, and is used to assess HCS’s ability to select key regions.
Notably, we partition the datasets to focus on the changes in the zero-shot transfer performance of VLMs after introducing HCS to evaluate its effectiveness.

\textbf{Implementation Details.}
HCS is implemented with a three-layer MLP that can be applied to any VLMs. It directly refines the interpretable regions based on the feature map of the baselines. Its integration allows the baselines, i.e., VLMs, to be tested without fine-tuning, instead training this lightweight network for coreset selection to enhance model performance. For optimization, we use the Adam optimizer \cite{kingma2014adam} with a momentum value of 0.8 and a weight decay of \(10^{-4}\). The initial learning rate is set to 0.1 and can be linearly scaled if necessary. We set $\lambda_{re}=0.7$, $\lambda_{ro}=0.65$, and $\lambda_{sy}=0.8$ via parameter sensitivity experiments. All the experiments are conducted over five runs on NVIDIA Tesla V100 GPUs.

\textbf{Evaluation Metrics.}
The experiments cover two main tasks: scene image classification and semantic segmentation. For image classification, accuracy is used as the evaluation metric, while for semantic segmentation, the Dice score is employed.

\subsection{Performance on Image Classification}
\label{sec:ex_2_performance_classification}
For scene image classification, we select multiple baselines to evaluate the zero-shot transfer performance, including ViT-B/32 \cite{liu2021swin}, ViT-L/14 \cite{liu2021swin}, CLIP \cite{CLIP}, ContextCLIP \cite{grover2022contextclip}, LLaVA-hf/llava-v1.6-mistral-7b-hf \cite{liu2023visual}, LLaVA-hf/llama3-llava-next-8b-hf \cite{liu2023visual}, and Qwen/Qwen2-VL-7B-Instruct \cite{bai2025qwen25vltechnicalreport}.
We adopt the weights of baseline VLMs that pre-trained on Conceptual Captions 1M (CC1M) \cite{sharma2018conceptual} image-caption pairs and Caltech256 \cite{griffin2007caltech} following \cite{li2021supervision,carlini2021poisoning,wang2024causal}. This configuration simulates real-world scenarios where existing pre-trained weights (published on Hugging Face etc.) are directly fine-tuned for downstream tasks. Additionally, we use concatenated wide-area remote sensing images as test data to evaluate the image classification performance before and after the introduction of HCS.

The results are shown in \textbf{Table \ref{tab:classification_results}}. From the results, we can observe that introducing HCS can achieve significant performance improvements across different baselines. Among them, for baseline models with relatively small parameter counts (e.g., ViT-B/32, ViT-L/14, CLIP, and ContextCLIP), Top-1 accuracy improved by an average of 15–18\% and Top-5 accuracy by above 15\%; for larger models (such as the LLaVA and Qwen series) also exhibited performance gains. These results indicate that HCS effectively refines image region interpretations, thereby substantially enhancing the models' zero-shot transfer generalization. Furthermore, given that HCS utilizes a lightweight three-layer MLP, it achieves significant performance improvements without requiring fine-tuning of the VLMs, further underscoring its practical superiority.

\subsection{Performance on Semantic Segmentation}
\label{sec:ex_3_performance_segmentation}
We also conduct comparison experiments on semantic segmentation using the three benchmark datasets mentioned in \textbf{Subsection \ref{sec:ex_1_settings}}, i.e., TikTok dances, TrashCan, and GTEA. 
Semantic segmentation, a core task in scene understanding, involves grouping pixels into regions with semantic classes. We mainly choose to compare with the SAM-based baselines, where SAM is proven to exhibit great performance in segmentation. The selected comparison baselines include Vanilla SAM \cite{sam}, Med-SA \cite{wu2023medical}, SAMed \cite{tancik2020fourier}, BLO-SAM \cite{zhangblo}, OVSAM \cite{yuan2024open}, and CLIPSAM \cite{CLIP}. To evaluate the transfer learning performance, each dataset is split into annotated base and unannotated target classes in a $1:1$ ratio, with fine-tuning based on the base class annotations from the pre-trained weights of SAM. Finally, we record the dice score on the unseen target classes.  

The results are shown in \textbf{Table \ref{tab:segmentation_results}}. For each SAM-based baseline, adding HCS consistently improves the Dice score on unseen target classes across all the benchmark datasets. For instance, Vanilla SAM's performance increases by at least 3.8\% on all the datasets; Similar improvements are observed for other SAM-based baselines, with gains ranging from approximately 1.5\% to 6.7\%. These results demonstrate the advantages of HCS, which effectively enhances segmentation performance across different models and datasets. 

\begin{figure*}
    \begin{minipage}[t]{0.39\textwidth}
        \centering
        \begin{subfigure}[t]{0.48\linewidth}
        \centering
        \includegraphics[width=\textwidth]{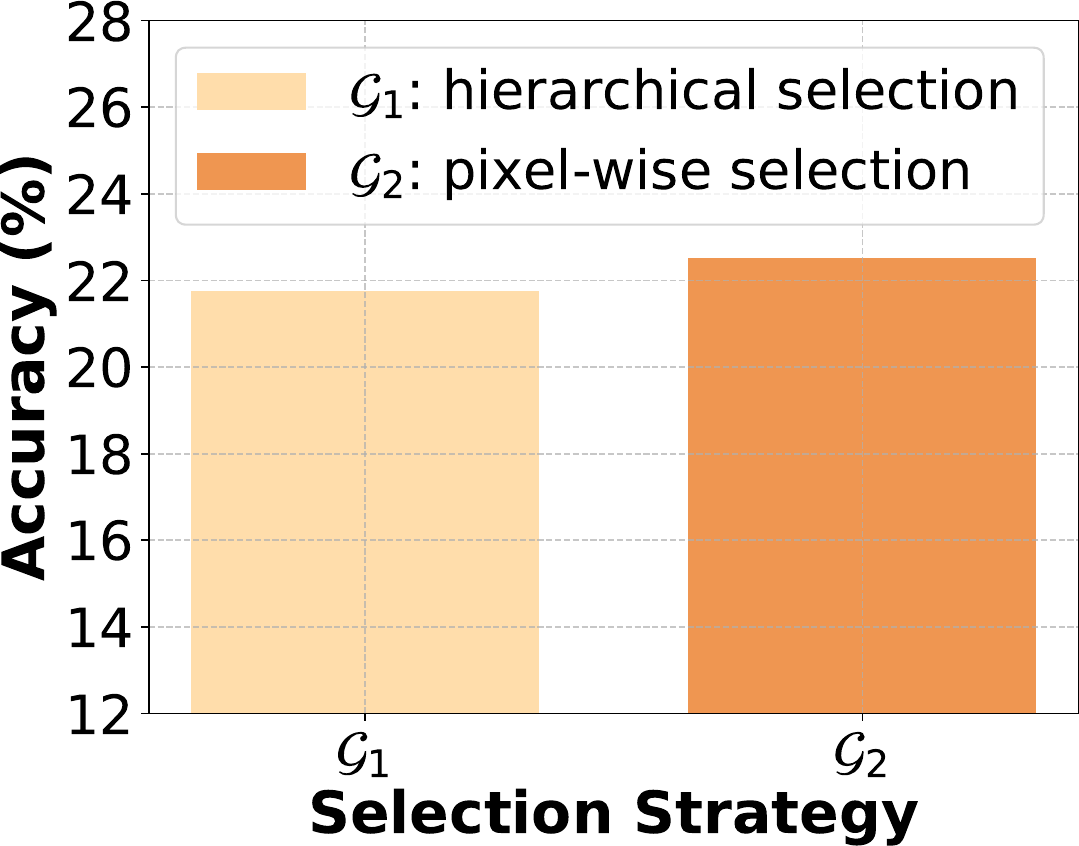}
        \caption{Performance} 
        \label{fig:abla_2_performance}
        \vspace{-0.1in}
    \end{subfigure}
    \hfill
    \begin{subfigure}[t]{0.48\linewidth}
        \centering
        \includegraphics[width=\textwidth]{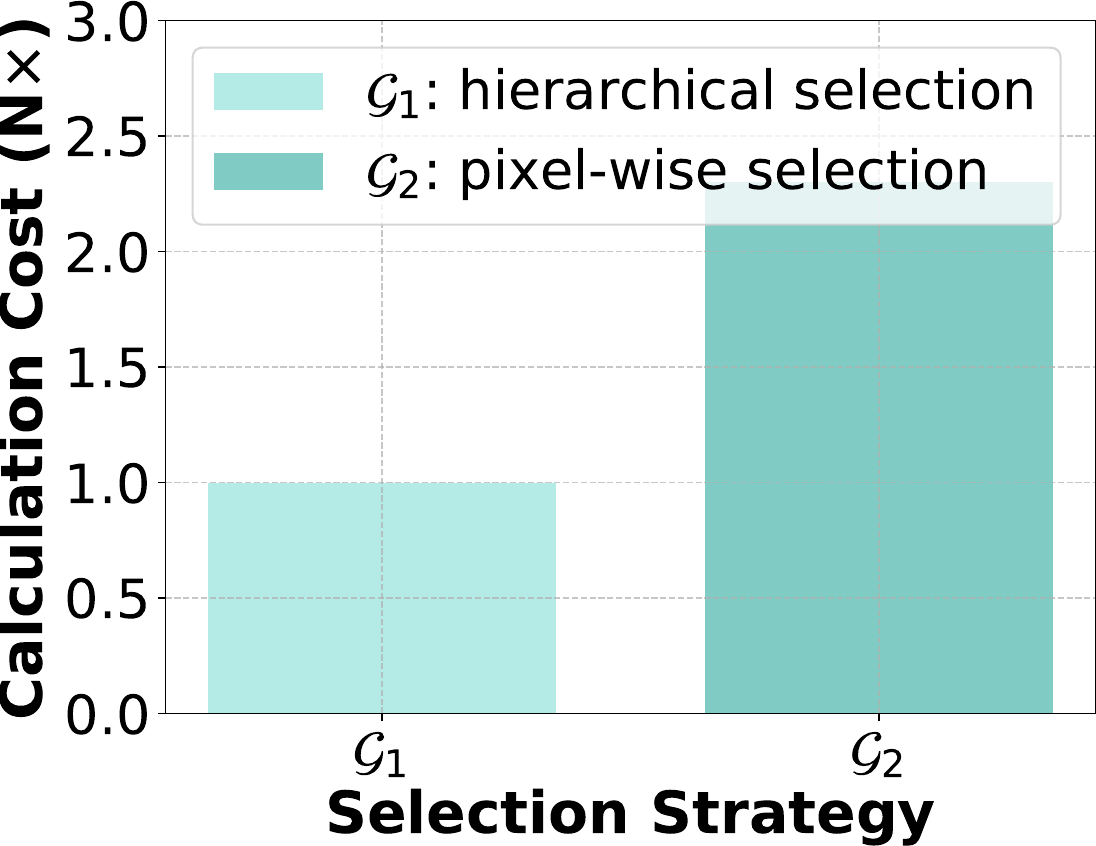}
        \caption{Efficiency} 
        \label{fig:abla_2_efficiency}
        \vspace{-0.1in}
    \end{subfigure}
    \caption{Impact of selection mechanism.}
    \label{fig:abla_2}
    \end{minipage}
    \hfill
    \begin{minipage}[t]{0.59\textwidth}
        \centering    
        \begin{subfigure}[t]{0.32\linewidth}
        \centering
        \includegraphics[width=\textwidth]{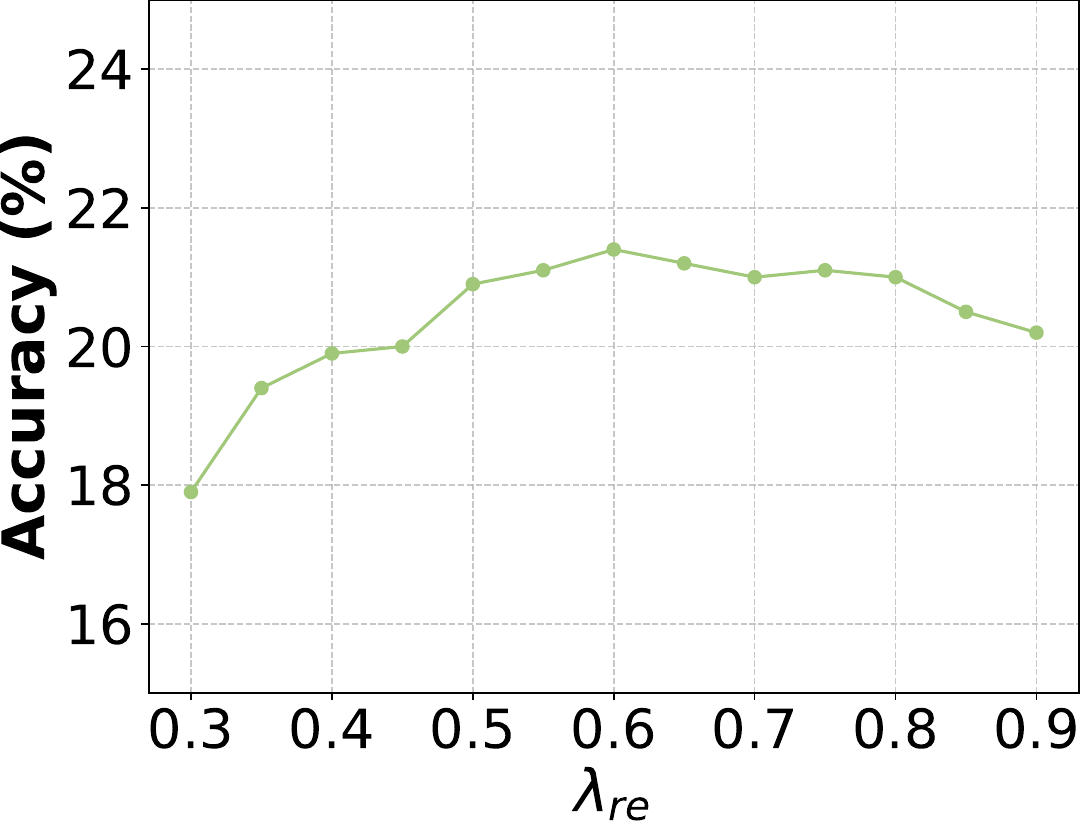}
        \caption{Results for $\lambda_{re}$} 
        \label{fig:abla_3_re}
        \vspace{-0.1in}
    \end{subfigure}
    \hfill
    \begin{subfigure}[t]{0.32\linewidth}
        \centering
        \includegraphics[width=\textwidth]{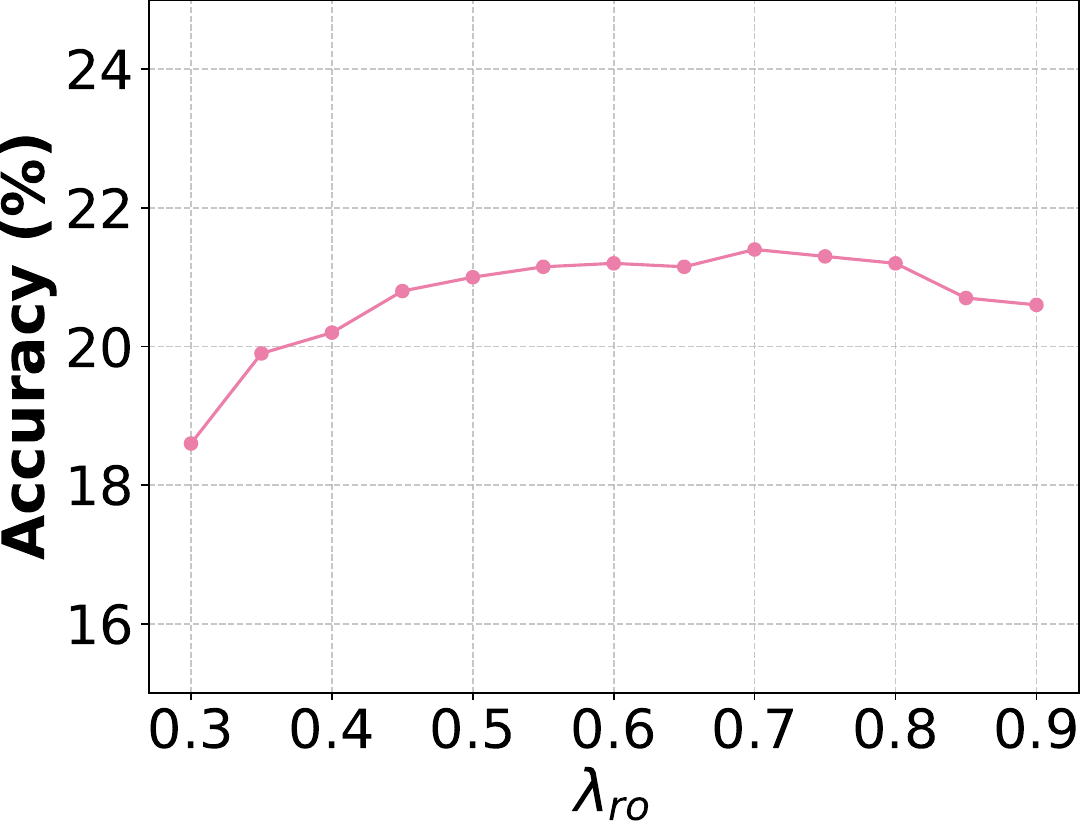}
        \caption{Results for $\lambda_{ro}$} 
        \label{fig:abla_3_ro}
        \vspace{-0.1in}
    \end{subfigure}
    \hfill
    \begin{subfigure}[t]{0.32\linewidth}
        \centering
        \includegraphics[width=\textwidth]{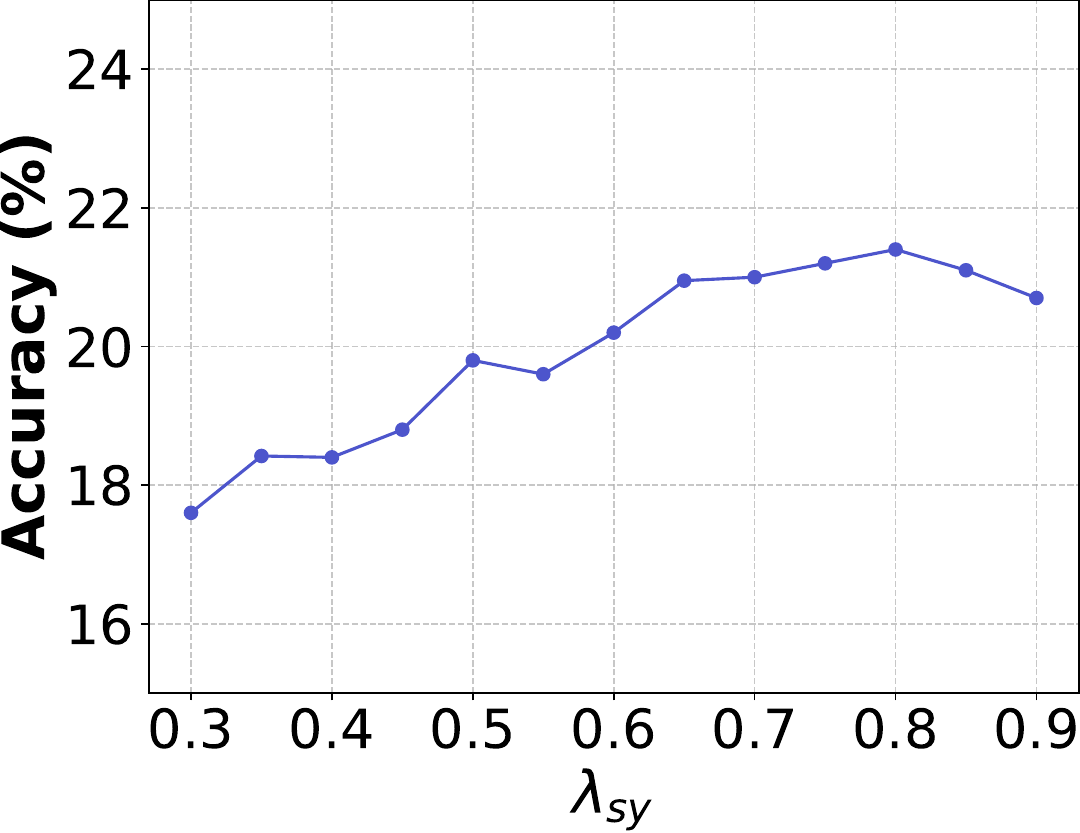}
        \caption{Results for $\lambda_{sy}$}
        \label{fig:abla_3_sy}
        \vspace{-0.1in}
    \end{subfigure}
    \caption{Ablation study of parameter sensitivity.}
    \label{fig:abla_3}
    \end{minipage}
\end{figure*}

\begin{figure*}
\begin{subfigure}{0.66\linewidth}
        \centering
        \includegraphics[width=\textwidth]{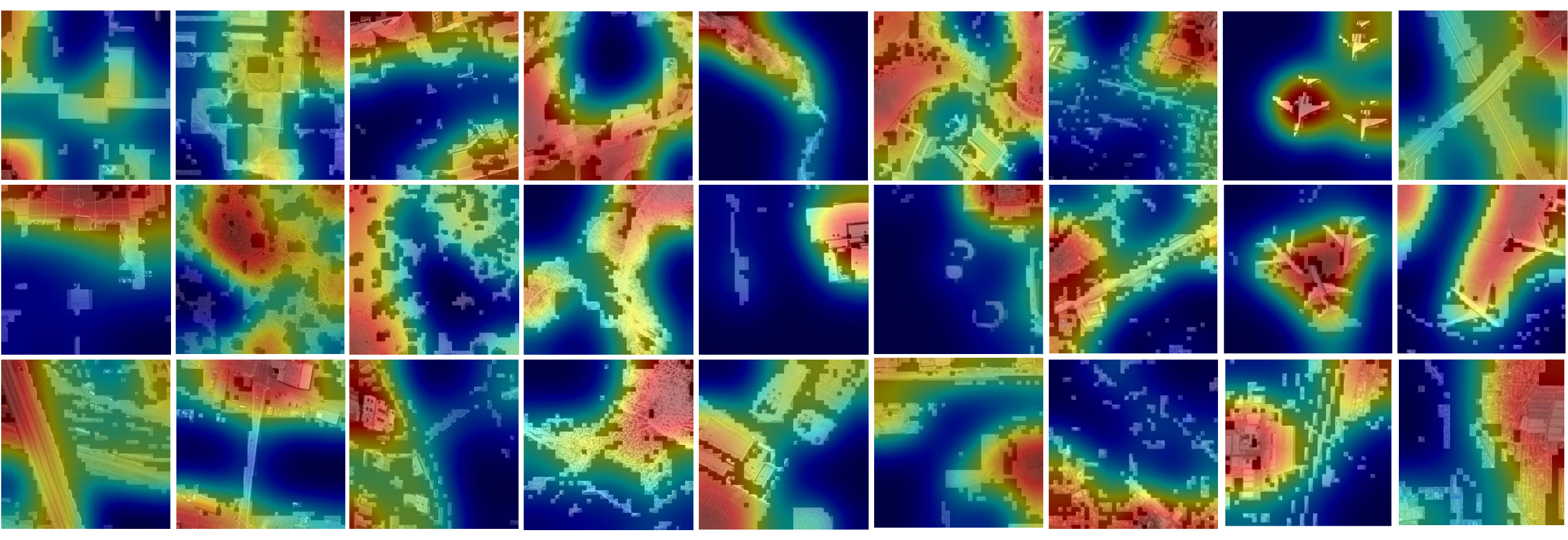}
        \caption{Visualization of interpretable regions}
        \vspace{-0.1in}
    \end{subfigure}
    \hfill
    \begin{subfigure}{0.3\linewidth}
        \centering
        \includegraphics[width=\textwidth]{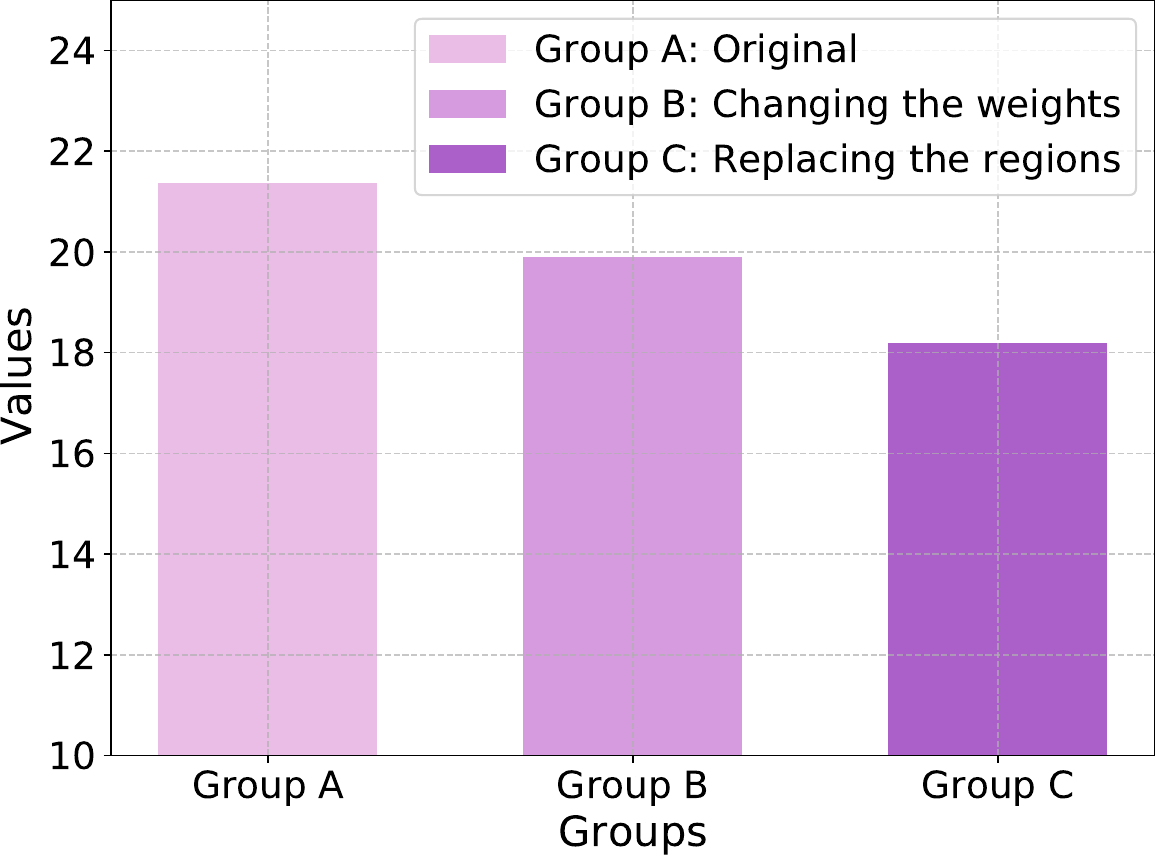}
        \caption{Performance comparison} 
        \vspace{-0.1in}
    \end{subfigure}
    \caption{Visualization results.  (a) shows the important regions identified by HCS. The unmasked areas represent the final selected interpretable regions, while the heatmap visualizes their corresponding importance scores. (b) shows the performance variation of CLIP+HCS under region perturbations: either by reducing the weights of elements above the median to 70\% of their original values or by randomly replacing 10\% of the selected regions with masked areas.}
    \label{fig:vis_heatmap}
    \vspace{-0.03in}
\end{figure*}

\subsection{Ablation Study}
\label{sec:ex_4_ablation}
In this subsection, we provide the results and analyses of the ablation studies, with more experiments in \textbf{Appendix \ref{sec_app:experiment}}.

\textbf{Impact of Different Components in Importance Function.} 
The key to the coreset selection of HCS lies in computing region scores using an importance function. As shown in Eq.\ref{eq:score}, the design of the importance function consists of the corresponding four components, i.e., $S_{ut}$, $S_{re}$, $S_{ro}$, and $S_{sy}$. To evaluate the impact of each component, we conduct ablation experiments on NWPU-RESISC45 with CLIP. We employ the same experimental settings as in \textbf{Subsection \ref{sec:ex_2_performance_classification}} and record the model performance after removing the corresponding component. Note that considering the definition of coreset (as illustrated in \textbf{Subsection \ref{sec:problem_settings}}), we always retain $S_{ut}$. The ablation results are shown in \textbf{Table \ref{tab:abla_1}}. From the results, we observe that all modules are important and contribute to performance improvements. This demonstrates the effectiveness of our design.

\textbf{Impact of Selection Mechanism.}
HCS employs a hierarchical search to ensure the coreset accuracy while reducing the computational cost (\textbf{Subsection \ref{sec:method_selection}}). To evaluate its effectiveness, we conduct an ablation experiment, i.e., compare the selection strategy with pixel-wise search. The pixel-wise search is considered to refine the smallest and accurate regions, but it may cause huge computational overhead. Using CLIP and taking the cost of HCS as the benchmark ($1\times$), we record the performance of both methods on NWPU-RESISC45. \textbf{Figure \ref{fig:abla_2}} shows that the computational cost of HCS is less than half that of the pixel-wise search, while the accuracy is nearly identical, demonstrating its superiority.

\textbf{Parameter Sensitivity.}
 We set the hyperparameters for the terms within the importance function (Eq.\ref{eq:score}), i.e., $\lambda_{re}$, $\lambda_{ro}$, and $\lambda_{sy}$ based on validation performance. For each experimental setting, we evaluate various values of $\lambda_{\cdot}$ within the range of $\left[0.3, 0.9 \right]$. For each hyperparameter $\lambda_{\cdot}$, initially, we perform a grid search with a step size of 0.05 to identify a promising interval. Then, within that interval, we refine our search with a step size of 0.01 and recorded the final average results. \textbf{Figure \ref{fig:abla_3}} shows that the best performance is achieved when $\lambda_{re}=0.7$, $\lambda_{ro}=0.65$, and $\lambda_{sy}=0.8$. Additionally, the model performance remains largely consistent across different parameter settings, demonstrating both the stability of the hyperparameter and the ease of tuning in practice. For simplicity, these parameters are set to 1 by default in our experiments (with a performance loss of no more than 2\%).

 \textbf{More Experiments and Results.} Note that the full analyses and more experiments regarding model efficiency and noise robustness, etc., are detailed in \textbf{Appendix \ref{sec_app:experiment}} due to space limitations.

\subsection{Visualization of Interpretable Regions}
\label{sec:ex_5_ablation}
To further elucidate the practical guidance provided by HCS and to show how the selected coresets reflect the most important regions in the samples, we construct more qualitative demonstrations for visualization.
Specifically, we randomly sample a set of samples from NWPU-RESISC45 and calculate the importance heatmap of the samples based on HCS. Meanwhile, we record the performance changes of (i) introducing HCS, and (ii) modifying scores calculated by HCS. The results are shown in \textbf{Figure \ref{fig:vis_heatmap}}. 
The results show that HCS selects coresets containing important semantics, such as foreground information, thereby improving the performance. Meanwhile, when we reduce the weights of high-weight elements and fed them directly into the frozen classification head, the accuracy decreased. This finding demonstrates the reliability of HCS.

\section{Conclusion}
\label{sec:conclusion}
In this paper, we address the challenge of wide-area scene understanding—a task that demands robust semantic extraction from complex, diverse, and sparsely distributed image regions. We reframe scene understanding as a coreset selection problem and propose a plug-and-play Hierarchical Coresets Selection (HCS) mechanism that refines region partitioning layer by layer. By leveraging the newly proposed importance function that considers utility, representativeness, robustness, and synergy, HCS accurately selects a small set of interpretable regions, enabling VLMs to achieve rapid, training-free scene understanding across different scales. Theoretical analyses demonstrate the superiority of the constructed importance function and the tighter generalization bound achieved by HCS. Extensive experiments on core scene understanding tasks, e.g., image classification and semantic segmentation, across a variety of VLM baselines, confirm that HCS consistently enhances model performance, particularly in challenging wide-area scenes.


\begin{acks}
The authors would like to thank the anonymous reviewers for their valuable comments. This work was supported in part by the Open Fund of (No. WDZC20255290403).
\end{acks}

\bibliographystyle{ACM-Reference-Format}
\bibliography{sample-base}


\begin{thebibliography}{76}


\ifx \showCODEN    \undefined \def \showCODEN     #1{\unskip}     \fi
\ifx \showISBNx    \undefined \def \showISBNx     #1{\unskip}     \fi
\ifx \showISBNxiii \undefined \def \showISBNxiii  #1{\unskip}     \fi
\ifx \showISSN     \undefined \def \showISSN      #1{\unskip}     \fi
\ifx \showLCCN     \undefined \def \showLCCN      #1{\unskip}     \fi
\ifx \shownote     \undefined \def \shownote      #1{#1}          \fi
\ifx \showarticletitle \undefined \def \showarticletitle #1{#1}   \fi
\ifx \showURL      \undefined \def \showURL       {\relax}        \fi
\providecommand\bibfield[2]{#2}
\providecommand\bibinfo[2]{#2}
\providecommand\natexlab[1]{#1}
\providecommand\showeprint[2][]{arXiv:#2}

\bibitem[Aarthi and Chitrakala(2017)]%
        {aarthi2017scene}
\bibfield{author}{\bibinfo{person}{S Aarthi} {and} \bibinfo{person}{S Chitrakala}.} \bibinfo{year}{2017}\natexlab{}.
\newblock \showarticletitle{Scene understanding—a survey}. In \bibinfo{booktitle}{\emph{2017 International conference on computer, communication and signal processing (ICCCSP)}}. IEEE, \bibinfo{pages}{1--4}.
\newblock


\bibitem[Asgari~Taghanaki et~al\mbox{.}(2021)]%
        {asgari2021deep}
\bibfield{author}{\bibinfo{person}{Saeid Asgari~Taghanaki}, \bibinfo{person}{Kumar Abhishek}, \bibinfo{person}{Joseph~Paul Cohen}, \bibinfo{person}{Julien Cohen-Adad}, {and} \bibinfo{person}{Ghassan Hamarneh}.} \bibinfo{year}{2021}\natexlab{}.
\newblock \showarticletitle{Deep semantic segmentation of natural and medical images: a review}.
\newblock \bibinfo{journal}{\emph{Artificial intelligence review}}  \bibinfo{volume}{54} (\bibinfo{year}{2021}), \bibinfo{pages}{137--178}.
\newblock


\bibitem[Ataee~Tarzanagh et~al\mbox{.}(2023)]%
        {ataee2023max}
\bibfield{author}{\bibinfo{person}{Davoud Ataee~Tarzanagh}, \bibinfo{person}{Yingcong Li}, \bibinfo{person}{Xuechen Zhang}, {and} \bibinfo{person}{Samet Oymak}.} \bibinfo{year}{2023}\natexlab{}.
\newblock \showarticletitle{Max-margin token selection in attention mechanism}.
\newblock \bibinfo{journal}{\emph{Advances in neural information processing systems}}  \bibinfo{volume}{36} (\bibinfo{year}{2023}), \bibinfo{pages}{48314--48362}.
\newblock


\bibitem[Bai et~al\mbox{.}(2025)]%
        {bai2025qwen25vltechnicalreport}
\bibfield{author}{\bibinfo{person}{Shuai Bai}, \bibinfo{person}{Keqin Chen}, \bibinfo{person}{Xuejing Liu}, \bibinfo{person}{Jialin Wang}, \bibinfo{person}{Wenbin Ge}, \bibinfo{person}{Sibo Song}, \bibinfo{person}{Kai Dang}, \bibinfo{person}{Peng Wang}, \bibinfo{person}{Shijie Wang}, \bibinfo{person}{Jun Tang}, \bibinfo{person}{Humen Zhong}, \bibinfo{person}{Yuanzhi Zhu}, \bibinfo{person}{Mingkun Yang}, \bibinfo{person}{Zhaohai Li}, \bibinfo{person}{Jianqiang Wan}, \bibinfo{person}{Pengfei Wang}, \bibinfo{person}{Wei Ding}, \bibinfo{person}{Zheren Fu}, \bibinfo{person}{Yiheng Xu}, \bibinfo{person}{Jiabo Ye}, \bibinfo{person}{Xi Zhang}, \bibinfo{person}{Tianbao Xie}, \bibinfo{person}{Zesen Cheng}, \bibinfo{person}{Hang Zhang}, \bibinfo{person}{Zhibo Yang}, \bibinfo{person}{Haiyang Xu}, {and} \bibinfo{person}{Junyang Lin}.} \bibinfo{year}{2025}\natexlab{}.
\newblock \bibinfo{title}{Qwen2.5-VL Technical Report}.
\newblock
\showeprint[arxiv]{2502.13923}~[cs.CV]
\urldef\tempurl%
\url{https://arxiv.org/abs/2502.13923}
\showURL{%
\tempurl}


\bibitem[Cao et~al\mbox{.}(2024)]%
        {cao2024maplm}
\bibfield{author}{\bibinfo{person}{Xu Cao}, \bibinfo{person}{Tong Zhou}, \bibinfo{person}{Yunsheng Ma}, \bibinfo{person}{Wenqian Ye}, \bibinfo{person}{Can Cui}, \bibinfo{person}{Kun Tang}, \bibinfo{person}{Zhipeng Cao}, \bibinfo{person}{Kaizhao Liang}, \bibinfo{person}{Ziran Wang}, \bibinfo{person}{James~M Rehg}, {et~al\mbox{.}}} \bibinfo{year}{2024}\natexlab{}.
\newblock \showarticletitle{Maplm: A real-world large-scale vision-language benchmark for map and traffic scene understanding}. In \bibinfo{booktitle}{\emph{Proceedings of the IEEE/CVF Conference on Computer Vision and Pattern Recognition}}. \bibinfo{pages}{21819--21830}.
\newblock


\bibitem[Carlini and Terzis(2021)]%
        {carlini2021poisoning}
\bibfield{author}{\bibinfo{person}{Nicholas Carlini} {and} \bibinfo{person}{Andreas Terzis}.} \bibinfo{year}{2021}\natexlab{}.
\newblock \showarticletitle{Poisoning and backdooring contrastive learning}.
\newblock \bibinfo{journal}{\emph{arXiv preprint arXiv:2106.09667}} (\bibinfo{year}{2021}).
\newblock


\bibitem[Chai et~al\mbox{.}(2023)]%
        {chai2023efficient}
\bibfield{author}{\bibinfo{person}{Chengliang Chai}, \bibinfo{person}{Jiayi Wang}, \bibinfo{person}{Nan Tang}, \bibinfo{person}{Ye Yuan}, \bibinfo{person}{Jiabin Liu}, \bibinfo{person}{Yuhao Deng}, {and} \bibinfo{person}{Guoren Wang}.} \bibinfo{year}{2023}\natexlab{}.
\newblock \showarticletitle{Efficient coreset selection with cluster-based methods}. In \bibinfo{booktitle}{\emph{Proceedings of the 29th ACM SIGKDD Conference on Knowledge Discovery and Data Mining}}. \bibinfo{pages}{167--178}.
\newblock


\bibitem[Chen et~al\mbox{.}(2023)]%
        {chen2023clip2scene}
\bibfield{author}{\bibinfo{person}{Runnan Chen}, \bibinfo{person}{Youquan Liu}, \bibinfo{person}{Lingdong Kong}, \bibinfo{person}{Xinge Zhu}, \bibinfo{person}{Yuexin Ma}, \bibinfo{person}{Yikang Li}, \bibinfo{person}{Yuenan Hou}, \bibinfo{person}{Yu Qiao}, {and} \bibinfo{person}{Wenping Wang}.} \bibinfo{year}{2023}\natexlab{}.
\newblock \showarticletitle{Clip2scene: Towards label-efficient 3d scene understanding by clip}. In \bibinfo{booktitle}{\emph{Proceedings of the IEEE/CVF Conference on Computer Vision and Pattern Recognition}}. \bibinfo{pages}{7020--7030}.
\newblock


\bibitem[Chen et~al\mbox{.}(2025)]%
        {chen2025advancinggeneralmultimodalcapability}
\bibfield{author}{\bibinfo{person}{Zhanpeng Chen}, \bibinfo{person}{Mingxiao Li}, \bibinfo{person}{Ziyang Chen}, \bibinfo{person}{Nan Du}, \bibinfo{person}{Xiaolong Li}, {and} \bibinfo{person}{Yuexian Zou}.} \bibinfo{year}{2025}\natexlab{}.
\newblock \bibinfo{title}{Advancing General Multimodal Capability of Vision-language Models with Pyramid-descent Visual Position Encoding}.
\newblock
\showeprint[arxiv]{2501.10967}~[cs.CV]
\urldef\tempurl%
\url{https://arxiv.org/abs/2501.10967}
\showURL{%
\tempurl}


\bibitem[Cheng et~al\mbox{.}(2017)]%
        {NWPU-RESISC45}
\bibfield{author}{\bibinfo{person}{Gong Cheng}, \bibinfo{person}{Junwei Han}, {and} \bibinfo{person}{Xiaoqiang Lu}.} \bibinfo{year}{2017}\natexlab{}.
\newblock \showarticletitle{Remote sensing image scene classification: Benchmark and state of the art}.
\newblock \bibinfo{journal}{\emph{Proc. IEEE}} \bibinfo{volume}{105}, \bibinfo{number}{10} (\bibinfo{year}{2017}), \bibinfo{pages}{1865--1883}.
\newblock


\bibitem[Cordts et~al\mbox{.}(2016)]%
        {cordts2016cityscapes}
\bibfield{author}{\bibinfo{person}{Marius Cordts}, \bibinfo{person}{Mohamed Omran}, \bibinfo{person}{Sebastian Ramos}, \bibinfo{person}{Timo Rehfeld}, \bibinfo{person}{Markus Enzweiler}, \bibinfo{person}{Rodrigo Benenson}, \bibinfo{person}{Uwe Franke}, \bibinfo{person}{Stefan Roth}, {and} \bibinfo{person}{Bernt Schiele}.} \bibinfo{year}{2016}\natexlab{}.
\newblock \showarticletitle{The cityscapes dataset for semantic urban scene understanding}. In \bibinfo{booktitle}{\emph{Proceedings of the IEEE conference on computer vision and pattern recognition}}. \bibinfo{pages}{3213--3223}.
\newblock


\bibitem[Drew et~al\mbox{.}(2013)]%
        {drew2013informatics}
\bibfield{author}{\bibinfo{person}{Trafton Drew}, \bibinfo{person}{Karla Evans}, \bibinfo{person}{Melissa L-H V{\~o}}, \bibinfo{person}{Francine~L Jacobson}, {and} \bibinfo{person}{Jeremy~M Wolfe}.} \bibinfo{year}{2013}\natexlab{}.
\newblock \showarticletitle{Informatics in radiology: what can you see in a single glance and how might this guide visual search in medical images?}
\newblock \bibinfo{journal}{\emph{Radiographics}} \bibinfo{volume}{33}, \bibinfo{number}{1} (\bibinfo{year}{2013}), \bibinfo{pages}{263--274}.
\newblock


\bibitem[Dubey et~al\mbox{.}(2015)]%
        {dubey2015coresetbasedadaptivetracking}
\bibfield{author}{\bibinfo{person}{Abhimanyu Dubey}, \bibinfo{person}{Nikhil Naik}, \bibinfo{person}{Dan Raviv}, \bibinfo{person}{Rahul Sukthankar}, {and} \bibinfo{person}{Ramesh Raskar}.} \bibinfo{year}{2015}\natexlab{}.
\newblock \bibinfo{title}{Coreset-Based Adaptive Tracking}.
\newblock
\showeprint[arxiv]{1511.06147}~[cs.CV]
\urldef\tempurl%
\url{https://arxiv.org/abs/1511.06147}
\showURL{%
\tempurl}


\bibitem[Fathi et~al\mbox{.}(2011)]%
        {fathi2011learning}
\bibfield{author}{\bibinfo{person}{Alireza Fathi}, \bibinfo{person}{Xiaofeng Ren}, {and} \bibinfo{person}{James~M Rehg}.} \bibinfo{year}{2011}\natexlab{}.
\newblock \showarticletitle{Learning to recognize objects in egocentric activities}. In \bibinfo{booktitle}{\emph{CVPR 2011}}. IEEE, \bibinfo{pages}{3281--3288}.
\newblock


\bibitem[Fu et~al\mbox{.}(2024)]%
        {fu2024scene}
\bibfield{author}{\bibinfo{person}{Rao Fu}, \bibinfo{person}{Jingyu Liu}, \bibinfo{person}{Xilun Chen}, \bibinfo{person}{Yixin Nie}, {and} \bibinfo{person}{Wenhan Xiong}.} \bibinfo{year}{2024}\natexlab{}.
\newblock \showarticletitle{Scene-llm: Extending language model for 3d visual understanding and reasoning}.
\newblock \bibinfo{journal}{\emph{arXiv preprint arXiv:2403.11401}} (\bibinfo{year}{2024}).
\newblock


\bibitem[Grant and Flynn(2017)]%
        {grant2017crowd}
\bibfield{author}{\bibinfo{person}{Jason~M Grant} {and} \bibinfo{person}{Patrick~J Flynn}.} \bibinfo{year}{2017}\natexlab{}.
\newblock \showarticletitle{Crowd scene understanding from video: a survey}.
\newblock \bibinfo{journal}{\emph{ACM Transactions on Multimedia Computing, Communications, and Applications (TOMM)}} \bibinfo{volume}{13}, \bibinfo{number}{2} (\bibinfo{year}{2017}), \bibinfo{pages}{1--23}.
\newblock


\bibitem[Greene and Oliva(2009)]%
        {greene2009briefest}
\bibfield{author}{\bibinfo{person}{Michelle~R Greene} {and} \bibinfo{person}{Aude Oliva}.} \bibinfo{year}{2009}\natexlab{}.
\newblock \showarticletitle{The briefest of glances: The time course of natural scene understanding}.
\newblock \bibinfo{journal}{\emph{Psychological science}} \bibinfo{volume}{20}, \bibinfo{number}{4} (\bibinfo{year}{2009}), \bibinfo{pages}{464--472}.
\newblock


\bibitem[Griffin et~al\mbox{.}(2007)]%
        {griffin2007caltech}
\bibfield{author}{\bibinfo{person}{Gregory Griffin}, \bibinfo{person}{Alex Holub}, \bibinfo{person}{Pietro Perona}, {et~al\mbox{.}}} \bibinfo{year}{2007}\natexlab{}.
\newblock \bibinfo{booktitle}{\emph{Caltech-256 object category dataset}}.
\newblock \bibinfo{type}{{T}echnical {R}eport}. \bibinfo{institution}{Technical Report 7694, California Institute of Technology Pasadena}.
\newblock


\bibitem[Grover et~al\mbox{.}(2022)]%
        {grover2022contextclip}
\bibfield{author}{\bibinfo{person}{Chanda Grover}, \bibinfo{person}{Indra~Deep Mastan}, {and} \bibinfo{person}{Debayan Gupta}.} \bibinfo{year}{2022}\natexlab{}.
\newblock \showarticletitle{Contextclip: Contextual alignment of image-text pairs on clip visual representations}. In \bibinfo{booktitle}{\emph{Proceedings of the Thirteenth Indian Conference on Computer Vision, Graphics and Image Processing}}. \bibinfo{pages}{1--10}.
\newblock


\bibitem[Gu et~al\mbox{.}(2019)]%
        {gu2019survey}
\bibfield{author}{\bibinfo{person}{Yating Gu}, \bibinfo{person}{Yantian Wang}, {and} \bibinfo{person}{Yansheng Li}.} \bibinfo{year}{2019}\natexlab{}.
\newblock \showarticletitle{A survey on deep learning-driven remote sensing image scene understanding: Scene classification, scene retrieval and scene-guided object detection}.
\newblock \bibinfo{journal}{\emph{Applied sciences}} \bibinfo{volume}{9}, \bibinfo{number}{10} (\bibinfo{year}{2019}), \bibinfo{pages}{2110}.
\newblock


\bibitem[Har-Peled and Mazumdar(2004)]%
        {har2004coresets}
\bibfield{author}{\bibinfo{person}{Sariel Har-Peled} {and} \bibinfo{person}{Soham Mazumdar}.} \bibinfo{year}{2004}\natexlab{}.
\newblock \showarticletitle{On coresets for k-means and k-median clustering}. In \bibinfo{booktitle}{\emph{Proceedings of the thirty-sixth annual ACM symposium on Theory of computing}}. \bibinfo{pages}{291--300}.
\newblock


\bibitem[Har-Peled and Sharir(2010)]%
        {RelativeApproximations}
\bibfield{author}{\bibinfo{person}{Sariel Har-Peled} {and} \bibinfo{person}{Micha Sharir}.} \bibinfo{year}{2010}\natexlab{}.
\newblock \bibinfo{title}{Relative $(p,\epsilon)$-Approximations in Geometry}.
\newblock
\showeprint[arxiv]{0909.0717}~[cs.CG]
\urldef\tempurl%
\url{https://arxiv.org/abs/0909.0717}
\showURL{%
\tempurl}


\bibitem[Hong et~al\mbox{.}(2020)]%
        {hong2020trashcan}
\bibfield{author}{\bibinfo{person}{Jungseok Hong}, \bibinfo{person}{Michael Fulton}, {and} \bibinfo{person}{Junaed Sattar}.} \bibinfo{year}{2020}\natexlab{}.
\newblock \showarticletitle{Trashcan: A semantically-segmented dataset towards visual detection of marine debris}.
\newblock \bibinfo{journal}{\emph{arXiv preprint arXiv:2007.08097}} (\bibinfo{year}{2020}).
\newblock


\bibitem[Hong et~al\mbox{.}(2018)]%
        {hong2018d3}
\bibfield{author}{\bibinfo{person}{Sungeun Hong}, \bibinfo{person}{Jongbin Ryu}, \bibinfo{person}{Woobin Im}, {and} \bibinfo{person}{Hyun~S Yang}.} \bibinfo{year}{2018}\natexlab{}.
\newblock \showarticletitle{D3: recognizing dynamic scenes with deep dual descriptor based on key frames and key segments}.
\newblock \bibinfo{journal}{\emph{Neurocomputing}}  \bibinfo{volume}{273} (\bibinfo{year}{2018}), \bibinfo{pages}{611--621}.
\newblock


\bibitem[Huang et~al\mbox{.}(2024)]%
        {huang2024optimal}
\bibfield{author}{\bibinfo{person}{Lingxiao Huang}, \bibinfo{person}{Jian Li}, {and} \bibinfo{person}{Xuan Wu}.} \bibinfo{year}{2024}\natexlab{}.
\newblock \showarticletitle{On optimal coreset construction for euclidean (k, z)-clustering}. In \bibinfo{booktitle}{\emph{Proceedings of the 56th Annual ACM Symposium on Theory of Computing}}. \bibinfo{pages}{1594--1604}.
\newblock


\bibitem[Huang et~al\mbox{.}(2018)]%
        {huang2018long}
\bibfield{author}{\bibinfo{person}{Yuanjun Huang}, \bibinfo{person}{Xianbin Cao}, \bibinfo{person}{Qi Wang}, \bibinfo{person}{Baochang Zhang}, \bibinfo{person}{Xiantong Zhen}, {and} \bibinfo{person}{Xuelong Li}.} \bibinfo{year}{2018}\natexlab{}.
\newblock \showarticletitle{Long-short-term features for dynamic scene classification}.
\newblock \bibinfo{journal}{\emph{IEEE Transactions on Circuits and Systems for Video Technology}} \bibinfo{volume}{29}, \bibinfo{number}{4} (\bibinfo{year}{2018}), \bibinfo{pages}{1038--1047}.
\newblock


\bibitem[Ibrahim(2016)]%
        {ibrahim2016comprehensive}
\bibfield{author}{\bibinfo{person}{Sutrisno~Warsono Ibrahim}.} \bibinfo{year}{2016}\natexlab{}.
\newblock \showarticletitle{A comprehensive review on intelligent surveillance systems}.
\newblock \bibinfo{journal}{\emph{Communications in science and technology}} \bibinfo{volume}{1}, \bibinfo{number}{1} (\bibinfo{year}{2016}).
\newblock


\bibitem[Jiang et~al\mbox{.}(2023)]%
        {jiang2023vad}
\bibfield{author}{\bibinfo{person}{Bo Jiang}, \bibinfo{person}{Shaoyu Chen}, \bibinfo{person}{Qing Xu}, \bibinfo{person}{Bencheng Liao}, \bibinfo{person}{Jiajie Chen}, \bibinfo{person}{Helong Zhou}, \bibinfo{person}{Qian Zhang}, \bibinfo{person}{Wenyu Liu}, \bibinfo{person}{Chang Huang}, {and} \bibinfo{person}{Xinggang Wang}.} \bibinfo{year}{2023}\natexlab{}.
\newblock \showarticletitle{Vad: Vectorized scene representation for efficient autonomous driving}. In \bibinfo{booktitle}{\emph{Proceedings of the IEEE/CVF International Conference on Computer Vision}}. \bibinfo{pages}{8340--8350}.
\newblock


\bibitem[Kingma and Ba(2014)]%
        {kingma2014adam}
\bibfield{author}{\bibinfo{person}{Diederik~P Kingma} {and} \bibinfo{person}{Jimmy Ba}.} \bibinfo{year}{2014}\natexlab{}.
\newblock \showarticletitle{Adam: A method for stochastic optimization}.
\newblock \bibinfo{journal}{\emph{arXiv preprint arXiv:1412.6980}} (\bibinfo{year}{2014}).
\newblock


\bibitem[Kirillov et~al\mbox{.}(2023)]%
        {sam}
\bibfield{author}{\bibinfo{person}{Alexander Kirillov}, \bibinfo{person}{Eric Mintun}, \bibinfo{person}{Nikhila Ravi}, \bibinfo{person}{Hanzi Mao}, \bibinfo{person}{Chloe Rolland}, \bibinfo{person}{Laura Gustafson}, \bibinfo{person}{Tete Xiao}, \bibinfo{person}{Spencer Whitehead}, \bibinfo{person}{Alexander~C Berg}, \bibinfo{person}{Wan-Yen Lo}, {et~al\mbox{.}}} \bibinfo{year}{2023}\natexlab{}.
\newblock \showarticletitle{Segment anything}. In \bibinfo{booktitle}{\emph{Proceedings of the IEEE/CVF International Conference on Computer Vision}}. \bibinfo{pages}{4015--4026}.
\newblock


\bibitem[Li et~al\mbox{.}(2020)]%
        {RSI-CB}
\bibfield{author}{\bibinfo{person}{Haifeng Li}, \bibinfo{person}{Xin Dou}, \bibinfo{person}{Chao Tao}, \bibinfo{person}{Zhixiang Wu}, \bibinfo{person}{Jie Chen}, \bibinfo{person}{Jian Peng}, \bibinfo{person}{Min Deng}, {and} \bibinfo{person}{Ling Zhao}.} \bibinfo{year}{2020}\natexlab{}.
\newblock \showarticletitle{RSI-CB: A Large-Scale Remote Sensing Image Classification Benchmark Using Crowdsourced Data}.
\newblock \bibinfo{journal}{\emph{Sensors}} \bibinfo{volume}{20}, \bibinfo{number}{6} (\bibinfo{year}{2020}).
\newblock
\showISSN{1424-8220}
\href{https://doi.org/10.3390/s20061594}{doi:\nolinkurl{10.3390/s20061594}}


\bibitem[Li et~al\mbox{.}(2009)]%
        {li2009towards}
\bibfield{author}{\bibinfo{person}{Li-Jia Li}, \bibinfo{person}{Richard Socher}, {and} \bibinfo{person}{Li Fei-Fei}.} \bibinfo{year}{2009}\natexlab{}.
\newblock \showarticletitle{Towards total scene understanding: Classification, annotation and segmentation in an automatic framework}. In \bibinfo{booktitle}{\emph{2009 IEEE Conference on Computer Vision and Pattern Recognition}}. IEEE, \bibinfo{pages}{2036--2043}.
\newblock


\bibitem[Li et~al\mbox{.}(2024)]%
        {li2024clipsam}
\bibfield{author}{\bibinfo{person}{Shengze Li}, \bibinfo{person}{Jianjian Cao}, \bibinfo{person}{Peng Ye}, \bibinfo{person}{Yuhan Ding}, \bibinfo{person}{Chongjun Tu}, {and} \bibinfo{person}{Tao Chen}.} \bibinfo{year}{2024}\natexlab{}.
\newblock \showarticletitle{ClipSAM: CLIP and SAM Collaboration for Zero-Shot Anomaly Segmentation}.
\newblock \bibinfo{journal}{\emph{arXiv preprint arXiv:2401.12665}} (\bibinfo{year}{2024}).
\newblock


\bibitem[Li et~al\mbox{.}(2021)]%
        {li2021supervision}
\bibfield{author}{\bibinfo{person}{Yangguang Li}, \bibinfo{person}{Feng Liang}, \bibinfo{person}{Lichen Zhao}, \bibinfo{person}{Yufeng Cui}, \bibinfo{person}{Wanli Ouyang}, \bibinfo{person}{Jing Shao}, \bibinfo{person}{Fengwei Yu}, {and} \bibinfo{person}{Junjie Yan}.} \bibinfo{year}{2021}\natexlab{}.
\newblock \showarticletitle{Supervision exists everywhere: A data efficient contrastive language-image pre-training paradigm}.
\newblock \bibinfo{journal}{\emph{arXiv preprint arXiv:2110.05208}} (\bibinfo{year}{2021}).
\newblock


\bibitem[Li et~al\mbox{.}(2001)]%
        {LI2001516}
\bibfield{author}{\bibinfo{person}{Yi Li}, \bibinfo{person}{Philip~M. Long}, {and} \bibinfo{person}{Aravind Srinivasan}.} \bibinfo{year}{2001}\natexlab{}.
\newblock \showarticletitle{Improved Bounds on the Sample Complexity of Learning}.
\newblock \bibinfo{journal}{\emph{J. Comput. System Sci.}} \bibinfo{volume}{62}, \bibinfo{number}{3} (\bibinfo{year}{2001}), \bibinfo{pages}{516--527}.
\newblock
\showISSN{0022-0000}
\href{https://doi.org/10.1006/jcss.2000.1741}{doi:\nolinkurl{10.1006/jcss.2000.1741}}


\bibitem[Liao et~al\mbox{.}(2024)]%
        {liao2024vlm2scene}
\bibfield{author}{\bibinfo{person}{Guibiao Liao}, \bibinfo{person}{Jiankun Li}, {and} \bibinfo{person}{Xiaoqing Ye}.} \bibinfo{year}{2024}\natexlab{}.
\newblock \showarticletitle{VLM2Scene: Self-supervised image-text-LiDAR learning with foundation models for autonomous driving scene understanding}. In \bibinfo{booktitle}{\emph{Proceedings of the AAAI Conference on Artificial Intelligence}}, Vol.~\bibinfo{volume}{38}. \bibinfo{pages}{3351--3359}.
\newblock


\bibitem[Liu et~al\mbox{.}(2023)]%
        {liu2023visual}
\bibfield{author}{\bibinfo{person}{Haotian Liu}, \bibinfo{person}{Chunyuan Li}, \bibinfo{person}{Qingyang Wu}, {and} \bibinfo{person}{Yong~Jae Lee}.} \bibinfo{year}{2023}\natexlab{}.
\newblock \showarticletitle{Visual instruction tuning}.
\newblock \bibinfo{journal}{\emph{Advances in neural information processing systems}}  \bibinfo{volume}{36} (\bibinfo{year}{2023}), \bibinfo{pages}{34892--34916}.
\newblock


\bibitem[Liu et~al\mbox{.}(2024)]%
        {liu2024vision}
\bibfield{author}{\bibinfo{person}{Sichao Liu}, \bibinfo{person}{Jianjing Zhang}, \bibinfo{person}{Robert~X Gao}, \bibinfo{person}{Xi~Vincent Wang}, {and} \bibinfo{person}{Lihui Wang}.} \bibinfo{year}{2024}\natexlab{}.
\newblock \showarticletitle{Vision-language model-driven scene understanding and robotic object manipulation}. In \bibinfo{booktitle}{\emph{2024 IEEE 20th International Conference on Automation Science and Engineering (CASE)}}. IEEE, \bibinfo{pages}{21--26}.
\newblock


\bibitem[Liu et~al\mbox{.}(2021)]%
        {liu2021swin}
\bibfield{author}{\bibinfo{person}{Ze Liu}, \bibinfo{person}{Yutong Lin}, \bibinfo{person}{Yue Cao}, \bibinfo{person}{Han Hu}, \bibinfo{person}{Yixuan Wei}, \bibinfo{person}{Zheng Zhang}, \bibinfo{person}{Stephen Lin}, {and} \bibinfo{person}{Baining Guo}.} \bibinfo{year}{2021}\natexlab{}.
\newblock \showarticletitle{Swin transformer: Hierarchical vision transformer using shifted windows}. In \bibinfo{booktitle}{\emph{Proceedings of the IEEE/CVF international conference on computer vision}}. \bibinfo{pages}{10012--10022}.
\newblock


\bibitem[Naseer et~al\mbox{.}(2018)]%
        {naseer2018indoor}
\bibfield{author}{\bibinfo{person}{Muzammal Naseer}, \bibinfo{person}{Salman Khan}, {and} \bibinfo{person}{Fatih Porikli}.} \bibinfo{year}{2018}\natexlab{}.
\newblock \showarticletitle{Indoor scene understanding in 2.5/3d for autonomous agents: A survey}.
\newblock \bibinfo{journal}{\emph{IEEE access}}  \bibinfo{volume}{7} (\bibinfo{year}{2018}), \bibinfo{pages}{1859--1887}.
\newblock


\bibitem[Neven et~al\mbox{.}(2017)]%
        {neven2017fast}
\bibfield{author}{\bibinfo{person}{Davy Neven}, \bibinfo{person}{Bert De~Brabandere}, \bibinfo{person}{Stamatios Georgoulis}, \bibinfo{person}{Marc Proesmans}, {and} \bibinfo{person}{Luc Van~Gool}.} \bibinfo{year}{2017}\natexlab{}.
\newblock \showarticletitle{Fast scene understanding for autonomous driving}.
\newblock \bibinfo{journal}{\emph{arXiv preprint arXiv:1708.02550}} (\bibinfo{year}{2017}).
\newblock


\bibitem[Peng and Bouzerdoum(2019)]%
        {peng2019part}
\bibfield{author}{\bibinfo{person}{Xiaoming Peng} {and} \bibinfo{person}{Abdesselam Bouzerdoum}.} \bibinfo{year}{2019}\natexlab{}.
\newblock \showarticletitle{Part-based feature aggregation method for dynamic scene recognition}. In \bibinfo{booktitle}{\emph{2019 Digital Image Computing: Techniques and Applications (DICTA)}}. IEEE, \bibinfo{pages}{1--8}.
\newblock


\bibitem[Phillips(2017)]%
        {phillips2017coresets}
\bibfield{author}{\bibinfo{person}{Jeff~M Phillips}.} \bibinfo{year}{2017}\natexlab{}.
\newblock \showarticletitle{Coresets and sketches}.
\newblock In \bibinfo{booktitle}{\emph{Handbook of discrete and computational geometry}}. \bibinfo{publisher}{Chapman and Hall/CRC}, \bibinfo{pages}{1269--1288}.
\newblock


\bibitem[Qi et~al\mbox{.}(2016)]%
        {qi2016dynamic}
\bibfield{author}{\bibinfo{person}{Xianbiao Qi}, \bibinfo{person}{Chun-Guang Li}, \bibinfo{person}{Guoying Zhao}, \bibinfo{person}{Xiaopeng Hong}, {and} \bibinfo{person}{Matti Pietik{\"a}inen}.} \bibinfo{year}{2016}\natexlab{}.
\newblock \showarticletitle{Dynamic texture and scene classification by transferring deep image features}.
\newblock \bibinfo{journal}{\emph{Neurocomputing}}  \bibinfo{volume}{171} (\bibinfo{year}{2016}), \bibinfo{pages}{1230--1241}.
\newblock


\bibitem[Qi et~al\mbox{.}(2025)]%
        {qi2025gpt4scene}
\bibfield{author}{\bibinfo{person}{Zhangyang Qi}, \bibinfo{person}{Zhixiong Zhang}, \bibinfo{person}{Ye Fang}, \bibinfo{person}{Jiaqi Wang}, {and} \bibinfo{person}{Hengshuang Zhao}.} \bibinfo{year}{2025}\natexlab{}.
\newblock \showarticletitle{GPT4Scene: Understand 3D Scenes from Videos with Vision-Language Models}.
\newblock \bibinfo{journal}{\emph{arXiv preprint arXiv:2501.01428}} (\bibinfo{year}{2025}).
\newblock


\bibitem[Qiang et~al\mbox{.}(2024)]%
        {qiang2024universality}
\bibfield{author}{\bibinfo{person}{Wenwen Qiang}, \bibinfo{person}{Jingyao Wang}, \bibinfo{person}{Lingyu Si}, {and} \bibinfo{person}{Changwen Zheng}.} \bibinfo{year}{2024}\natexlab{}.
\newblock \showarticletitle{On the Universality of Self-Supervised Representation Learning}.
\newblock  (\bibinfo{year}{2024}).
\newblock


\bibitem[Radford et~al\mbox{.}(2021)]%
        {CLIP}
\bibfield{author}{\bibinfo{person}{Alec Radford}, \bibinfo{person}{Jong~Wook Kim}, \bibinfo{person}{Chris Hallacy}, \bibinfo{person}{Aditya Ramesh}, \bibinfo{person}{Gabriel Goh}, \bibinfo{person}{Sandhini Agarwal}, \bibinfo{person}{Girish Sastry}, \bibinfo{person}{Amanda Askell}, \bibinfo{person}{Pamela Mishkin}, \bibinfo{person}{Jack Clark}, \bibinfo{person}{Gretchen Krueger}, {and} \bibinfo{person}{Ilya Sutskever}.} \bibinfo{year}{2021}\natexlab{}.
\newblock \showarticletitle{Learning Transferable Visual Models From Natural Language Supervision}.
\newblock \bibinfo{journal}{\emph{CoRR}}  \bibinfo{volume}{abs/2103.00020} (\bibinfo{year}{2021}).
\newblock
\showeprint[arXiv]{2103.00020}
\urldef\tempurl%
\url{https://arxiv.org/abs/2103.00020}
\showURL{%
\tempurl}


\bibitem[Redmon and Farhadi(2018)]%
        {redmon2018yolov3}
\bibfield{author}{\bibinfo{person}{Joseph Redmon} {and} \bibinfo{person}{Ali Farhadi}.} \bibinfo{year}{2018}\natexlab{}.
\newblock \showarticletitle{Yolov3: An incremental improvement}.
\newblock \bibinfo{journal}{\emph{arXiv preprint arXiv:1804.02767}} (\bibinfo{year}{2018}).
\newblock


\bibitem[Roman(2023)]%
        {roman2023humantiktok}
\bibfield{author}{\bibinfo{person}{Karpovich Roman}.} \bibinfo{year}{2023}\natexlab{}.
\newblock \bibinfo{title}{Human Segmentation Dataset - TikTok Dances}.
\newblock \bibinfo{howpublished}{\url{https://www.kaggle.com/datasets}}.
\newblock


\bibitem[Sayood(2017)]%
        {sayood2017introduction}
\bibfield{author}{\bibinfo{person}{Khalid Sayood}.} \bibinfo{year}{2017}\natexlab{}.
\newblock \bibinfo{booktitle}{\emph{Introduction to data compression}}.
\newblock \bibinfo{publisher}{Morgan Kaufmann}.
\newblock


\bibitem[Sener and Savarese(2017)]%
        {sener2017active}
\bibfield{author}{\bibinfo{person}{Ozan Sener} {and} \bibinfo{person}{Silvio Savarese}.} \bibinfo{year}{2017}\natexlab{}.
\newblock \showarticletitle{Active learning for convolutional neural networks: A core-set approach}.
\newblock \bibinfo{journal}{\emph{arXiv preprint arXiv:1708.00489}} (\bibinfo{year}{2017}).
\newblock


\bibitem[Sharma et~al\mbox{.}(2018)]%
        {sharma2018conceptual}
\bibfield{author}{\bibinfo{person}{Piyush Sharma}, \bibinfo{person}{Nan Ding}, \bibinfo{person}{Sebastian Goodman}, {and} \bibinfo{person}{Radu Soricut}.} \bibinfo{year}{2018}\natexlab{}.
\newblock \showarticletitle{Conceptual captions: A cleaned, hypernymed, image alt-text dataset for automatic image captioning}. In \bibinfo{booktitle}{\emph{Proceedings of the 56th Annual Meeting of the Association for Computational Linguistics (Volume 1: Long Papers)}}. \bibinfo{pages}{2556--2565}.
\newblock


\bibitem[Shi et~al\mbox{.}(2024)]%
        {shi2024vila}
\bibfield{author}{\bibinfo{person}{Jiangbo Shi}, \bibinfo{person}{Chen Li}, \bibinfo{person}{Tieliang Gong}, \bibinfo{person}{Yefeng Zheng}, {and} \bibinfo{person}{Huazhu Fu}.} \bibinfo{year}{2024}\natexlab{}.
\newblock \showarticletitle{ViLa-MIL: Dual-scale Vision-Language Multiple Instance Learning for Whole Slide Image Classification}. In \bibinfo{booktitle}{\emph{Proceedings of the IEEE/CVF Conference on Computer Vision and Pattern Recognition}}. \bibinfo{pages}{11248--11258}.
\newblock


\bibitem[Sohler and Woodruff(2022)]%
        {sohler2022strongcoresetskmediansubspace}
\bibfield{author}{\bibinfo{person}{Christian Sohler} {and} \bibinfo{person}{David~P. Woodruff}.} \bibinfo{year}{2022}\natexlab{}.
\newblock \bibinfo{title}{Strong Coresets for k-Median and Subspace Approximation: Goodbye Dimension}.
\newblock
\showeprint[arxiv]{1809.02961}~[cs.DS]
\urldef\tempurl%
\url{https://arxiv.org/abs/1809.02961}
\showURL{%
\tempurl}


\bibitem[Storer(1987)]%
        {storer1987data}
\bibfield{author}{\bibinfo{person}{James~A Storer}.} \bibinfo{year}{1987}\natexlab{}.
\newblock \bibinfo{booktitle}{\emph{Data compression: methods and theory}}.
\newblock \bibinfo{publisher}{Computer Science Press, Inc.}
\newblock


\bibitem[Tancik et~al\mbox{.}(2020)]%
        {tancik2020fourier}
\bibfield{author}{\bibinfo{person}{Matthew Tancik}, \bibinfo{person}{Pratul Srinivasan}, \bibinfo{person}{Ben Mildenhall}, \bibinfo{person}{Sara Fridovich-Keil}, \bibinfo{person}{Nithin Raghavan}, \bibinfo{person}{Utkarsh Singhal}, \bibinfo{person}{Ravi Ramamoorthi}, \bibinfo{person}{Jonathan Barron}, {and} \bibinfo{person}{Ren Ng}.} \bibinfo{year}{2020}\natexlab{}.
\newblock \showarticletitle{Fourier features let networks learn high frequency functions in low dimensional domains}.
\newblock \bibinfo{journal}{\emph{Advances in neural information processing systems}}  \bibinfo{volume}{33} (\bibinfo{year}{2020}), \bibinfo{pages}{7537--7547}.
\newblock


\bibitem[Tukan et~al\mbox{.}(2020)]%
        {tukan2020coresets}
\bibfield{author}{\bibinfo{person}{Murad Tukan}, \bibinfo{person}{Alaa Maalouf}, {and} \bibinfo{person}{Dan Feldman}.} \bibinfo{year}{2020}\natexlab{}.
\newblock \showarticletitle{Coresets for near-convex functions}.
\newblock \bibinfo{journal}{\emph{Advances in Neural Information Processing Systems}}  \bibinfo{volume}{33} (\bibinfo{year}{2020}), \bibinfo{pages}{997--1009}.
\newblock


\bibitem[Vepa et~al\mbox{.}(2025)]%
        {vepa2025integrating}
\bibfield{author}{\bibinfo{person}{Arvind Vepa}, \bibinfo{person}{Zukang Yang}, \bibinfo{person}{Andrew Choi}, \bibinfo{person}{Jungseock Joo}, \bibinfo{person}{Fabien Scalzo}, {and} \bibinfo{person}{Yizhou Sun}.} \bibinfo{year}{2025}\natexlab{}.
\newblock \showarticletitle{Integrating Deep Metric Learning with Coreset for Active Learning in 3D Segmentation}.
\newblock \bibinfo{journal}{\emph{Advances in Neural Information Processing Systems}}  \bibinfo{volume}{37} (\bibinfo{year}{2025}), \bibinfo{pages}{71643--71671}.
\newblock


\bibitem[Wang et~al\mbox{.}(2023a)]%
        {wang2023amsa}
\bibfield{author}{\bibinfo{person}{Jingyao Wang}, \bibinfo{person}{Luntian Mou}, \bibinfo{person}{Lei Ma}, \bibinfo{person}{Tiejun Huang}, {and} \bibinfo{person}{Wen Gao}.} \bibinfo{year}{2023}\natexlab{a}.
\newblock \showarticletitle{AMSA: Adaptive multimodal learning for sentiment analysis}.
\newblock \bibinfo{journal}{\emph{ACM Transactions on Multimedia Computing, Communications and Applications}} \bibinfo{volume}{19}, \bibinfo{number}{3s} (\bibinfo{year}{2023}), \bibinfo{pages}{1--21}.
\newblock


\bibitem[Wang et~al\mbox{.}(2024c)]%
        {wang2024image}
\bibfield{author}{\bibinfo{person}{Jingyao Wang}, \bibinfo{person}{Luntian Mou}, \bibinfo{person}{Changwen Zheng}, {and} \bibinfo{person}{Wen Gao}.} \bibinfo{year}{2024}\natexlab{c}.
\newblock \showarticletitle{Image-based Freeform Handwriting Authentication with Energy-oriented Self-Supervised Learning}.
\newblock \bibinfo{journal}{\emph{arXiv preprint arXiv:2408.09676}} (\bibinfo{year}{2024}).
\newblock


\bibitem[Wang et~al\mbox{.}(2024d)]%
        {wang2024causal}
\bibfield{author}{\bibinfo{person}{Jingyao Wang}, \bibinfo{person}{Wenwen Qiang}, \bibinfo{person}{Jiangmeng Li}, \bibinfo{person}{Lingyu Si}, \bibinfo{person}{Changwen Zheng}, {and} \bibinfo{person}{Bing Su}.} \bibinfo{year}{2024}\natexlab{d}.
\newblock \showarticletitle{On the Causal Sufficiency and Necessity of Multi-Modal Representation Learning}.
\newblock \bibinfo{journal}{\emph{arXiv preprint arXiv:2407.14058}} (\bibinfo{year}{2024}).
\newblock


\bibitem[Wang et~al\mbox{.}(2024e)]%
        {wang2024towards}
\bibfield{author}{\bibinfo{person}{Jingyao Wang}, \bibinfo{person}{Wenwen Qiang}, \bibinfo{person}{Xingzhe Su}, \bibinfo{person}{Changwen Zheng}, \bibinfo{person}{Fuchun Sun}, {and} \bibinfo{person}{Hui Xiong}.} \bibinfo{year}{2024}\natexlab{e}.
\newblock \showarticletitle{Towards Task Sampler Learning for Meta-Learning}.
\newblock \bibinfo{journal}{\emph{International Journal of Computer Vision}} (\bibinfo{year}{2024}), \bibinfo{pages}{1--31}.
\newblock


\bibitem[Wang and Yu(2022)]%
        {wang2022so}
\bibfield{author}{\bibinfo{person}{Jingyao Wang} {and} \bibinfo{person}{Naigong Yu}.} \bibinfo{year}{2022}\natexlab{}.
\newblock \showarticletitle{So-perm: Pose estimation and robust measurement for small objects}. In \bibinfo{booktitle}{\emph{2022 International Joint Conference on Neural Networks (IJCNN)}}. IEEE, \bibinfo{pages}{1--7}.
\newblock


\bibitem[Wang et~al\mbox{.}(2023b)]%
        {wang2023awesome}
\bibfield{author}{\bibinfo{person}{Jingyao Wang}, \bibinfo{person}{Chuyuan Zhang}, \bibinfo{person}{Ye Ding}, {and} \bibinfo{person}{Yuxuan Yang}.} \bibinfo{year}{2023}\natexlab{b}.
\newblock \showarticletitle{Awesome-META+: Meta-Learning Research and Learning Platform}.
\newblock \bibinfo{journal}{\emph{arXiv preprint arXiv:2304.12921}} (\bibinfo{year}{2023}).
\newblock


\bibitem[Wang et~al\mbox{.}(2024b)]%
        {wang2024tsp}
\bibfield{author}{\bibinfo{person}{Shuo Wang}, \bibinfo{person}{Jing Li}, \bibinfo{person}{Zibo Zhao}, \bibinfo{person}{Dongze Lian}, \bibinfo{person}{Binbin Huang}, \bibinfo{person}{Xiaomei Wang}, \bibinfo{person}{Zhengxin Li}, {and} \bibinfo{person}{Shenghua Gao}.} \bibinfo{year}{2024}\natexlab{b}.
\newblock \showarticletitle{Tsp-transformer: Task-specific prompts boosted transformer for holistic scene understanding}. In \bibinfo{booktitle}{\emph{Proceedings of the IEEE/CVF Winter Conference on Applications of Computer Vision}}. \bibinfo{pages}{925--934}.
\newblock


\bibitem[Wang et~al\mbox{.}(2024a)]%
        {wang2024data}
\bibfield{author}{\bibinfo{person}{Sheng-Yu Wang}, \bibinfo{person}{Aaron Hertzmann}, \bibinfo{person}{Alexei Efros}, \bibinfo{person}{Jun-Yan Zhu}, {and} \bibinfo{person}{Richard Zhang}.} \bibinfo{year}{2024}\natexlab{a}.
\newblock \showarticletitle{Data attribution for text-to-image models by unlearning synthesized images}.
\newblock \bibinfo{journal}{\emph{Advances in Neural Information Processing Systems}}  \bibinfo{volume}{37} (\bibinfo{year}{2024}), \bibinfo{pages}{4235--4266}.
\newblock


\bibitem[Wang(2024)]%
        {wang2024reasoning}
\bibfield{author}{\bibinfo{person}{Zhonghao Wang}.} \bibinfo{year}{2024}\natexlab{}.
\newblock \emph{\bibinfo{title}{Reasoning, scaling, generating with vision-language models}}.
\newblock \bibinfo{thesistype}{Ph.\,D. Dissertation}. \bibinfo{school}{University of Illinois at Urbana-Champaign}.
\newblock


\bibitem[Wu et~al\mbox{.}(2023)]%
        {wu2023medical}
\bibfield{author}{\bibinfo{person}{Junde Wu}, \bibinfo{person}{Wei Ji}, \bibinfo{person}{Yuanpei Liu}, \bibinfo{person}{Huazhu Fu}, \bibinfo{person}{Min Xu}, \bibinfo{person}{Yanwu Xu}, {and} \bibinfo{person}{Yueming Jin}.} \bibinfo{year}{2023}\natexlab{}.
\newblock \showarticletitle{Medical sam adapter: Adapting segment anything model for medical image segmentation}.
\newblock \bibinfo{journal}{\emph{arXiv preprint arXiv:2304.12620}} (\bibinfo{year}{2023}).
\newblock


\bibitem[Xia et~al\mbox{.}(2017)]%
        {AID}
\bibfield{author}{\bibinfo{person}{Gui-Song Xia}, \bibinfo{person}{Jingwen Hu}, \bibinfo{person}{Fan Hu}, \bibinfo{person}{Baoguang Shi}, \bibinfo{person}{Xiang Bai}, \bibinfo{person}{Yanfei Zhong}, \bibinfo{person}{Liangpei Zhang}, {and} \bibinfo{person}{Xiaoqiang Lu}.} \bibinfo{year}{2017}\natexlab{}.
\newblock \showarticletitle{AID: A Benchmark Data Set for Performance Evaluation of Aerial Scene Classification}.
\newblock \bibinfo{journal}{\emph{IEEE Transactions on Geoscience and Remote Sensing}} \bibinfo{volume}{55}, \bibinfo{number}{7} (\bibinfo{year}{2017}), \bibinfo{pages}{3965--3981}.
\newblock
\href{https://doi.org/10.1109/TGRS.2017.2685945}{doi:\nolinkurl{10.1109/TGRS.2017.2685945}}


\bibitem[Xie et~al\mbox{.}(2020)]%
        {xie2020pointcontrast}
\bibfield{author}{\bibinfo{person}{Saining Xie}, \bibinfo{person}{Jiatao Gu}, \bibinfo{person}{Demi Guo}, \bibinfo{person}{Charles~R Qi}, \bibinfo{person}{Leonidas Guibas}, {and} \bibinfo{person}{Or Litany}.} \bibinfo{year}{2020}\natexlab{}.
\newblock \showarticletitle{Pointcontrast: Unsupervised pre-training for 3d point cloud understanding}. In \bibinfo{booktitle}{\emph{Computer Vision--ECCV 2020: 16th European Conference, Glasgow, UK, August 23--28, 2020, Proceedings, Part III 16}}. Springer, \bibinfo{pages}{574--591}.
\newblock


\bibitem[Yuan et~al\mbox{.}(2024)]%
        {yuan2024open}
\bibfield{author}{\bibinfo{person}{Haobo Yuan}, \bibinfo{person}{Xiangtai Li}, \bibinfo{person}{Chong Zhou}, \bibinfo{person}{Yining Li}, \bibinfo{person}{Kai Chen}, {and} \bibinfo{person}{Chen~Change Loy}.} \bibinfo{year}{2024}\natexlab{}.
\newblock \showarticletitle{Open-vocabulary SAM: Segment and recognize twenty-thousand classes interactively}.
\newblock \bibinfo{journal}{\emph{arXiv preprint arXiv:2401.02955}} (\bibinfo{year}{2024}).
\newblock


\bibitem[Zhan et~al\mbox{.}(2025)]%
        {zhan2025coreset}
\bibfield{author}{\bibinfo{person}{Donglin Zhan}, \bibinfo{person}{Leonardo~F Toso}, {and} \bibinfo{person}{James Anderson}.} \bibinfo{year}{2025}\natexlab{}.
\newblock \showarticletitle{Coreset-Based Task Selection for Sample-Efficient Meta-Reinforcement Learning}.
\newblock \bibinfo{journal}{\emph{arXiv preprint arXiv:2502.02332}} (\bibinfo{year}{2025}).
\newblock


\bibitem[Zhang et~al\mbox{.}(2024b)]%
        {zhangblo}
\bibfield{author}{\bibinfo{person}{Li Zhang}, \bibinfo{person}{Youwei Liang}, \bibinfo{person}{Ruiyi Zhang}, \bibinfo{person}{Amirhosein Javadi}, {and} \bibinfo{person}{Pengtao Xie}.} \bibinfo{year}{2024}\natexlab{b}.
\newblock \showarticletitle{BLO-SAM: Bi-level Optimization Based Finetuning of the Segment Anything Model for Overfitting-Preventing Semantic Segmentation}. In \bibinfo{booktitle}{\emph{Forty-first International Conference on Machine Learning}}.
\newblock


\bibitem[Zhang et~al\mbox{.}(2024a)]%
        {zhang2024cls}
\bibfield{author}{\bibinfo{person}{Qizhe Zhang}, \bibinfo{person}{Aosong Cheng}, \bibinfo{person}{Ming Lu}, \bibinfo{person}{Zhiyong Zhuo}, \bibinfo{person}{Minqi Wang}, \bibinfo{person}{Jiajun Cao}, \bibinfo{person}{Shaobo Guo}, \bibinfo{person}{Qi She}, {and} \bibinfo{person}{Shanghang Zhang}.} \bibinfo{year}{2024}\natexlab{a}.
\newblock \showarticletitle{[CLS] Attention is All You Need for Training-Free Visual Token Pruning: Make VLM Inference Faster}.
\newblock \bibinfo{journal}{\emph{arXiv preprint arXiv:2412.01818}} (\bibinfo{year}{2024}).
\newblock


\bibitem[Zhang et~al\mbox{.}(2012)]%
        {zhang2012mining}
\bibfield{author}{\bibinfo{person}{Tianzhu Zhang}, \bibinfo{person}{Si Liu}, \bibinfo{person}{Changsheng Xu}, {and} \bibinfo{person}{Hanqing Lu}.} \bibinfo{year}{2012}\natexlab{}.
\newblock \showarticletitle{Mining semantic context information for intelligent video surveillance of traffic scenes}.
\newblock \bibinfo{journal}{\emph{IEEE transactions on industrial informatics}} \bibinfo{volume}{9}, \bibinfo{number}{1} (\bibinfo{year}{2012}), \bibinfo{pages}{149--160}.
\newblock


\bibitem[Zhou et~al\mbox{.}(2024)]%
        {zhou2024embodied}
\bibfield{author}{\bibinfo{person}{Yunsong Zhou}, \bibinfo{person}{Linyan Huang}, \bibinfo{person}{Qingwen Bu}, \bibinfo{person}{Jia Zeng}, \bibinfo{person}{Tianyu Li}, \bibinfo{person}{Hang Qiu}, \bibinfo{person}{Hongzi Zhu}, \bibinfo{person}{Minyi Guo}, \bibinfo{person}{Yu Qiao}, {and} \bibinfo{person}{Hongyang Li}.} \bibinfo{year}{2024}\natexlab{}.
\newblock \showarticletitle{Embodied understanding of driving scenarios}. In \bibinfo{booktitle}{\emph{European Conference on Computer Vision}}. Springer, \bibinfo{pages}{129--148}.
\newblock


\end{thebibliography}

\clearpage
\newpage
\appendix

\section*{Appendix}
This supplementary material provides results for additional experiments and details to reproduce our results that could not be included in the paper submission due to space limitations.
\begin{itemize}
    \item \textbf{Appendix \ref{sec_app:proof}} provides proofs and further theoretical analysis of the theory in the text.
    \item \textbf{Appendix \ref{sec_app:dataset}} provides the additional details of the benchmark datasets.
    \item \textbf{Appendix \ref{sec_app:baselines}} provides the additional details of the baselines for comparision.
    \item \textbf{Appendix \ref{sec_app:implementation}} provides the additional details of the implementation details.
    \item \textbf{Appendix \ref{sec_app:experiment}} provides the full results and additional experiments. 
\end{itemize}

Note that before we illustrate the details and analysis, we provide a brief summary of all the experiments conducted in this paper and a summary of notations, as shown in Table \ref{tab:app} and Table \ref{tab:notations}.

\begin{table*}[htpb]
    \centering
    \caption{Illustration of the experiments conducted in this work. Note that all experimental results are obtained after five rounds of experiments.}
    \begin{tabular}{p{0.4\textwidth}|p{0.3\textwidth}|p{0.2\textwidth}}
    \toprule
        \textbf{Experiments} & \textbf{Location} & \textbf{Results}\\
    \midrule    
        Performance comparison of scene image classification & Section \ref{sec:ex_2_performance_classification} & Table \ref{tab:classification_results}\\
    \midrule    
        Performance comparison of semantic segmentation & Section \ref{sec:ex_3_performance_segmentation} & Table \ref{tab:segmentation_results}\\
    \midrule    
        Ablation study about the impact of different components in importance function & Section \ref{sec:ex_4_ablation} & Table \ref{tab:abla_1} \\
    \midrule    
        Ablation study about the impact of selection mechanism & Section \ref{sec:ex_4_ablation} & Figure \ref{fig:abla_2} \\
    \midrule
        Experiment of parameter sensitivity & Section \ref{sec:ex_4_ablation} and Appendix \ref{sec_app:parameter} & Figure \ref{fig:abla_3} and Figure \ref{fig:app_parameter}\\
    \midrule  
        Experiment of model efficiency & Appendix \ref{sec_app:experiment_trade-off} & Figure \ref{fig:ex_model_effi}  \\    
    \midrule
        Performance of scene understanding when facing noise & Appendix \ref{sec_app:experiment_noise} & Table \ref{tab:app_noise}\\
    \bottomrule
    \end{tabular}
    \label{tab:app}
\end{table*}

\begin{table*}[t]
    \centering
    \caption{The definitions of notations.}
    \label{tab:notations}
    \begin{tabular}{c|c}
    \toprule
    Notations & Definition \\
    \midrule
    \multicolumn{2}{c}{Notations of Data and Model} \\
    \midrule
    $X$ & The input wide-area image $X\in \mathbb{R}^{H \times W \times C}$ with height \(H\), width \(W\), and \(C\) channels\\
    \(Y\) & The ground truth of $X$ (e.g., semantic labels or detection boxes)\\
    \((x, y) \in \mathcal{X} \times \mathcal{Y}\) &  The decomposed units \(x \in \mathcal{X}\) and the label \(y \in \mathcal{Y}\) \\
    \(\Theta\) & The parameter space of candidate models\\
    \(P\) & The probability measure over \(\mathcal{X}\)\\
    $f_\theta: X \rightarrow Y$ & The target model, i.e., VLMs for scene understanding\\
    \(\ell: \mathcal{X} \times \Theta \to [0,\infty)\) & The loss function defined on a single unit \(x\)\\
    $\mathcal{L}(\theta) = \int_{x \in \mathcal{X}} \ell(x,\theta) \, \mathrm{d}P(x)$ & The full loss evaluated at a model parameter \(\theta \in \Theta\)\\
    \midrule
    \multicolumn{2}{c}{Notations of Coreset and HCS} \\
    \midrule
    $X_s$ & The $\epsilon$-coreset of $X$, i.e., $\left| \mathcal{L}(\theta) - \sum_{x \in X} \nu(x) \, \ell(x,\theta) \right| \le \epsilon\, \mathcal{L}(\theta)$ \\
    \(\nu: X \to \mathbb{R}_{\ge 0}\) & The weight function \\
    \(\epsilon \in (0,1)\) & The tolerance of $X_s$\\
    $S_{ut}(\cdot)$ & Score of utility (Eq.\ref{eq:score_ut}) \\
    $S_{re}(\cdot)$ & Score of representativeness (Eq.\ref{eq:score_re}) \\
    $S_{ro}(\cdot)$ & Score of robustness (Eq.\ref{eq:score_ro}) \\
    $S_{sy}(\cdot)$ & Score of synergy (Eq.\ref{eq:score_sy}) \\
    $S(\cdot)$ & Score of importance (Eq.\ref{eq:score}), i.e., $S(X_s)=S_{ut}(X_s)+\lambda_{re} S_{re}(X_s)+ \lambda_{ro}S_{ro}(X_s)+\lambda_{sy} S_{sy}(X_s)$ \\
    $\lambda_{re}$, $\lambda_{ro}$, $\lambda_{sy}$ & The weight of $S_{re}(\cdot)$, $S_{ro}(\cdot)$, and $S_{sy}(\cdot)$\\
    \( \mathcal{R} = \{r_{ij}\} \) & The set of regions $r_{ij}$, $i,j=1,\dots,N$\\
    \( \mathcal{R}_s \) & The candidate set determined by $S(\cdot)$, i.e., \( X_s = \bigcup_{r \in \mathcal{R}_{s}} r \) \\
    \midrule
    \multicolumn{2}{c}{Notations of Theoretical Analysis} \\
    \midrule
    $S$ & The total importance score for coreset $X_s$, i.e., $S = \int_{\mathcal{X}} s(x)\,dP(x)$\\
    $s:\mathcal{X}\to (1,\infty)$ & The importance function with for \(\ell(\cdot,\theta)\) with total score $S$ \\
    $m$ & The size of selected coreset, $m\ge g(\epsilon,S,\delta)=\frac{2S}{\epsilon^2}\Bigl(8C^2+S\ln\frac{2}{\delta}\Bigr)$ \\
    \(\epsilon\) & The tolerance \\
    \(\delta\) & The confidence parameter\\
    $g(\epsilon,S,\delta)$ & The function for size $m$\\
    $\mathcal{H}$ & The hypothesis class of the model, $\mathcal{H}\subseteq\{h:\mathbb{R}^d\to \mathcal{Y}\}$\\
     $H$ & The pseudo-dimension, $H = \mathrm{Pdim}(\{\ell_h : h\in\mathcal{H}\})$\\
     \(\mathfrak{R}(\mathcal{H})\) & The Rademacher complexity of \(\mathcal{H}\)\\
     \(R(f_\theta,X)\) & The risk on the full image \(X\)\\
     \(\widehat{R}(f_\theta, X_s)\) & The empirical risk computed on the coreset \(X_s\)\\
     \(M\) & A finite constant\\
    \bottomrule
    \end{tabular}
\end{table*}

\section{Proofs}
\label{sec_app:proof}
In this section, we prove two main results: (i) \textbf{Theorem \ref{theo:1}}, which establishes that the coreset selected by our HCS method is both bounded and effective for wide-area scene understanding while also achieving a reduction in computational complexity; and (ii) \textbf{Theorem \ref{theo:2}}, which demonstrates that the coreset chosen using HCS not only theoretically approximates the full loss but, when combined with standard generalization error bounds derived from Rademacher complexity and pseudo-dimension, also results in a tighter overall upper bound.

\subsection{Proof of Theorem \ref{theo:1}}
\label{sec_app:proof_1}
\emph{Proof.}
Recall that our setting is as follows. Let
$X\in\mathbb{R}^{w\times h\times c}$ be a wide-area scene image and let \(\mathcal{X}\) denote the set of minimal units extracted from \(X\). The full loss of a fixed model \(f_\theta\) is defined by $\mathcal{L}(\theta)=\int_{\mathcal{X}}\ell(x,\theta)\,dP(x)$ where \(P\) is a probability measure on \(\mathcal{X}\). We assume there exists an upper importance function $s:\mathcal{X}\to (1,\infty)$ for \(\ell(\cdot,\theta)\) with total score $S=\int_{\mathcal{X}} s(x)\,dP(x)$ via our HCS procedure, we obtain a coreset \(X_s\subset\mathcal{X}\) of size \(m\) and assign each selected unit the weight $\nu(x)=\frac{S}{m\,s(x)}$.
Our goal is to show that, with probability at least \(1-\delta\), for all \(\theta\) the weighted loss estimate $\hat{\mathcal{L}}(\theta)=\sum_{x\in X_s} \nu(x)\,\ell(x,\theta)$ satisfies $\Big|\mathcal{L}(\theta)-\hat{\mathcal{L}}(\theta)\Big|\le\epsilon\,\mathcal{L}(\theta)$.

To that end, we begin by normalizing the loss. For any fixed \(\theta\), define the sensitivity‐normalized loss:  
\begin{equation}
A(\theta;x)=\frac{S\,\ell(x,\theta)}{s(x)\,\mathcal{L}(\theta)}.
\end{equation}
Then, by linearity we have:
\begin{equation}
\int_{\mathcal{X}} A(\theta;x)\,dP(x)
=\frac{S}{\mathcal{L}(\theta)}\int_{\mathcal{X}} \frac{\ell(x,\theta)}{s(x)}\,dP(x).
\end{equation}
However, it is more natural to work with the probability measure \(Q\) defined via $dQ(x)=\frac{s(x)}{S}\,dP(x)$.
Indeed, under \(Q\) the normalized loss becomes
\begin{equation}
\begin{split}
    \int_{\mathcal{X}} A(\theta;x)\,dQ(x)
    &=\frac{S}{\mathcal{L}(\theta)}\int_{\mathcal{X}} \frac{\ell(x,\theta)}{s(x)}\,dQ(x)\\
    &=\frac{1}{\mathcal{L}(\theta)}\int_{\mathcal{X}} \ell(x,\theta)\,dP(x)=1.
\end{split}
\end{equation}
Thus, for every \(\theta\) the expected value of \(A(\theta;x)\) under \(Q\) is 1.

The coreset \(X_s\) is obtained by drawing \(m\) i.i.d. samples from \(Q\) according to the sensitivity distribution. The corresponding empirical (normalized) loss is:
\begin{equation}
\widehat{A}(\theta)=\frac{1}{m}\sum_{x\in X_s}A(\theta;x)=\frac{1}{\mathcal{L}(\theta)}\sum_{x\in X_s} \frac{S\,\ell(x,\theta)}{m\,s(x)},
\end{equation}
so that
$\Big|\mathcal{L}(\theta)-\hat{\mathcal{L}}(\theta)\Big|=\mathcal{L}(\theta)\Big|1-\widehat{A}(\theta)\Big|$.
To guarantee a relative error bound of \(\epsilon\), it suffices to ensure that $\Big|1-\widehat{A}(\theta)\Big|\le \epsilon\quad\text{for all } \theta$.
In order to achieve this, we rely on the concept of a relative \((\epsilon,\eta)\)-approximation for range spaces. In our context, the family of coreset that is defined in \cite{har2004coresets} can be expressed as:
\begin{equation}
\mathcal{R}=\Big\{ \{x\in\mathcal{X}:A(\theta;x)>t\}:\, \theta\in\Theta,\; t\ge0 \Big\}
\end{equation}
forms a range space on \((\mathcal{X},Q)\) with bounded VC-dimension. By standard results following \cite{RelativeApproximations,LI2001516}, there exists a universal constant \(C\) such that, if we set an additive parameter \(\eta=1/S\), then an i.i.d. sample \(X_s\) of size $m\ge \frac{4CS}{\epsilon^2}\Bigl(\mathsf{VC}\log S+\ln\frac{1}{\delta}\Bigr)$
will, with probability at least \(1-\delta\), satisfy:
\begin{equation}
\left| Q(R) - \frac{|R\cap X_s|}{m}\right|\le \frac{\epsilon}{2}\max\Bigl(\frac{1}{S},\, Q(R)\Bigr)\quad\forall R\in \mathcal{R}.
\end{equation}
This means that the uniform empirical distribution over \(X_s\) is a relative \(\Bigl(\frac{\epsilon}{2},\frac{1}{S}\Bigr)\)-approximation for the range space induced by \(A(\theta;\cdot)\).

Following \cite{sohler2022strongcoresetskmediansubspace,dubey2015coresetbasedadaptivetracking}, we can obtain that if a measure \(\nu\) is a relative \((\epsilon,\eta)\)-approximation for the range space (with \(\eta=1/S\)), then for each \(\theta\) we have:
\begin{equation}
\Biggl|\int_{\mathcal{X}} A(\theta;x)\,dQ(x)-\int_{\mathcal{X}} A(\theta;x)\,d\nu(x)\Biggr|\le \epsilon\Biggl(1+S\cdot\frac{1}{S}\Biggr)=2\epsilon.
\end{equation}
After rescaling, this yields the unconditioned relative error $\Big|1-\widehat{A}(\theta)\Big|\le \epsilon$
(Any constant factors can be absorbed by adjusting the universal constant \(C\) in the sample size bound).
Finally, returning to the original loss, we have:  
\begin{equation}
\Big|\mathcal{L}(\theta)-\hat{\mathcal{L}}(\theta)\Big|=\mathcal{L}(\theta)\Big|1-\widehat{A}(\theta)\Big|\le \epsilon\,\mathcal{L}(\theta),
\end{equation}

Thus, by normalizing the loss with respect to an importance-weighted measure \(Q\) and applying VC-dimension–based relative approximation results (with \(\eta=1/S\)), we conclude that if  
$m\ge \frac{2S}{\epsilon^2}\Bigl(8C^2+S\ln\frac{2}{\delta}\Bigr)$,
then the weighted loss computed on the coreset \(X_s\) approximates the full loss \(\mathcal{L}(\theta)\) within a relative error of \(\epsilon\) for all \(\theta\), with probability at least \(1-\delta\). This completes the proof of Theorem \ref{theo:1}.

\subsection{Proof of Theorem \ref{theo:2}}
\label{sec_app:proof_2}
We begin by defining a modified margin function that naturally reflects the effect of coreset selection. For any \(h\in\mathcal{H}\) and any sample \((x,y)\), define
\begin{equation}
\varphi_{\lambda,h}(x,y)=\min_{\hat{y}\in\mathcal{Y}}\Bigl\{h(x,y)-h(x,\hat{y})+\lambda\,\mathbf{1}_{\{\hat{y}=y\}}\Bigr\},
\end{equation}
with \(\lambda>0\) an arbitrary constant. Note that when \(\hat{y}\neq y\) the indicator vanishes, so that
\begin{equation}
\varphi_{\lambda,h}(x,y) \le \min_{\hat{y}\neq y}\Bigl\{h(x,y)-h(x,\hat{y})\Bigr\}\triangleq \varphi_{0,h}(x,y).
\end{equation}
This immediately implies
\begin{equation}
\mathbf{1}\{\varphi_{0,h}(x,y)\le0\}\le \mathbf{1}\{\varphi_{\lambda,h}(x,y)\le0\},
\end{equation}
which will allow us to control the natural error indicator by a surrogate loss.

To facilitate the analysis, we introduce a smoothing function \(\psi_\lambda:\mathbb{R}\to\mathbb{R}\) that satisfies: (i) for every \(u\in\mathbb{R}\), \(\mathbf{1}\{u\le0\}\le \psi_\lambda(u)\), and (ii) \(\psi_\lambda\) is \(\tfrac{1}{\lambda}\)-Lipschitz (i.e., \(|\psi_\lambda(u)-\psi_\lambda(v)|\le \tfrac{1}{\lambda}|u-v|\) for all \(u,v\)). We then define the composite function
\begin{equation}
g_h(x,y)=\psi_\lambda\bigl(\varphi_{\lambda,h}(x,y)\bigr)
\end{equation}
and consider the function class
\begin{equation}
\mathcal{G}=\{g_h: h\in\mathcal{H}\}.
\end{equation}
Because of the pointwise inequality, the risk on the full image,
\begin{equation}
R(f_\theta,X)=\mathbb{E}\Bigl[\mathbf{1}\{\varphi_{0,h}(x,y)\le0\}\Bigr],
\end{equation}
is upper-bounded by the expected surrogate loss:
\begin{equation}
R(f_\theta,X)\le \mathbb{E}\Bigl[g_h(x,y)\Bigr].
\end{equation}

A key feature of our approach is that the coreset \(X_s\) selected via HCS satisfies
\begin{equation}
\left|\mathcal{L}(X)-\mathcal{L}(X_s)\right|\le \epsilon\,\mathcal{L}(X),
\end{equation}
which guarantees that the empirical risk computed over \(X_s\) is a close approximation of the full-image risk. In particular, if we define
\begin{equation}
\widehat{R}(f_\theta,X_s)=\frac{1}{m}\sum_{i=1}^m g_h(x_i,y_i),
\end{equation}
where \((x_i,y_i)\) are samples drawn from \(X_s\) (with \(m\) being the number of such samples), then the coreset property ensures that the empirical evaluation is reliable.

Using standard uniform convergence arguments based on Rademacher complexity—specifically, applying symmetrization and contraction inequalities—we obtain that with probability at least \(1-\delta\) over the draw of an \(m\)-sample from \(\mathcal{D}_{tr}\), the following inequality holds for all \(h\in\mathcal{H}\):
\begin{equation}
\mathbb{E}\Bigl[g_h(x,y)\Bigr] \le \frac{1}{m}\sum_{i=1}^m g_h(x_i,y_i) + 2\,\mathfrak{R}_m(\mathcal{G}) + \sqrt{\frac{\ln(1/\delta)}{2m}}.
\end{equation}
Thus, we have
\begin{equation}
R(f_\theta,X)\le \widehat{R}(f_\theta,X_s) + 2\,\mathfrak{R}_m(\mathcal{G}) + \sqrt{\frac{\ln(1/\delta)}{2m}}.
\end{equation}

We then choose the smoothing parameter by setting \(\lambda=2\eta\) for some \(\eta>0\). A careful case analysis (examining whether the minimum in the definition of \(\varphi_{\lambda,h}\) is achieved when \(\hat{y}=y\) or not) shows that with this choice the surrogate satisfies
\begin{equation}
\psi_\lambda\bigl(\varphi_{\lambda,h}(x_i,y_i)\bigr)=\psi_\lambda\bigl(\varphi_{0,h}(x_i,y_i)\bigr)
\end{equation}
for every sample \((x_i,y_i)\). In other words, the smoothed empirical risk \(\widehat{R}(f_\theta,X_s)\) effectively reflects the natural margin function.

It remains to bound the Rademacher complexity term \(\mathfrak{R}_m(\mathcal{G})\). By the contraction lemma, since \(\psi_\lambda\) is \(\tfrac{1}{\lambda}\)-Lipschitz, we have
\begin{equation}
\mathfrak{R}_m(\mathcal{G})\le \frac{1}{\lambda}\,\mathfrak{R}_m\Bigl(\bigl\{(x,y)\mapsto \varphi_{\lambda,h}(x,y):h\in\mathcal{H}\bigr\}\Bigr).
\end{equation}
For any sample \((x_i,y_i)\), note that
\begin{equation}
\varphi_{\lambda,h}(x_i,y_i)=h(x_i,y_i)-\max_{z\in\mathcal{Y}}\Bigl\{h(x_i,z)-\lambda\,\mathbf{1}_{\{z=y_i\}}\Bigr\}.
\end{equation}
Standard techniques allow us to decompose the Rademacher average as
\begin{equation}
\begin{aligned}
&\mathfrak{R}_m\Bigl(\{\varphi_{\lambda,h}\}\Bigr)\\
&=\frac{1}{m}\mathbb{E}_{S,\sigma}\left[\sup_{h\in\mathcal{H}}\sum_{i=1}^m\sigma_i\Bigl(h(x_i,y_i)-\max_{z\in\mathcal{Y}}\{h(x_i,z)-\lambda\,\mathbf{1}_{\{z=y_i\}}\}\Bigr)\right]\\
&\le \frac{1}{m}\mathbb{E}_{S,\sigma}\left[\sup_{h\in\mathcal{H}}\sum_{i=1}^m\sigma_i\,h(x_i,y_i)\right] \\
&\quad+\frac{1}{m}\mathbb{E}_{S,\sigma}\left[\sup_{h\in\mathcal{H}}\sum_{i=1}^m\sigma_i\,\max_{z\in\mathcal{Y}}\{h(x_i,z)-\lambda\,\mathbf{1}_{\{z=y_i\}}\}\right].
\end{aligned}
\end{equation}
For the first term, we express \(h(x_i,y_i)=\sum_{z\in\mathcal{Y}}h(x_i,z)\,\mathbf{1}_{\{z=y_i\}}\) and then apply linearity and symmetry of the Rademacher variables. This yields a bound proportional to a constant (depending on \(|\mathcal{Y}|\)) times \(\mathfrak{R}_m\bigl(\Pi_1(\mathcal{H})\bigr)\), where \(\Pi_1(\mathcal{H})\) denotes the projection of \(\mathcal{H}\) onto its coordinate functions. A similar bound applies to the second term. Therefore, there exists a constant \(C>0\) such that
\begin{equation}
\mathfrak{R}_m\Bigl(\{\varphi_{\lambda,h}\}\Bigr)\le C\,\mathfrak{R}_m\Bigl(\Pi_1(\mathcal{H})\Bigr).
\end{equation}
Thus, by the contraction inequality,
\begin{equation}
\mathfrak{R}_m(\mathcal{G})\le \frac{C}{\lambda}\,\mathfrak{R}_m\Bigl(\Pi_1(\mathcal{H})\Bigr).
\end{equation}
Choosing \(\lambda=2\eta\) and absorbing the constant factors into a finite constant \(M\), we deduce that
\begin{equation}
2\,\mathfrak{R}_m(\mathcal{G})\le M\,\mathfrak{R}(\mathcal{H}).
\end{equation}

Substituting this bound into our earlier inequality, we obtain
\begin{equation}
R(f_\theta,X)\le \widehat{R}(f_\theta,X_s)+M\,\mathfrak{R}(\mathcal{H})+\sqrt{\frac{\ln(1/\delta)}{2m}}.
\end{equation}
Finally, by relating the sample size \(m\) (which here reflects the effective number of coreset samples) to the pseudo-dimension \(H\) of the loss class, we can replace the deviation term \(\sqrt{\frac{\ln(1/\delta)}{2m}}\) with \(\sqrt{\frac{\ln(1/\delta)}{2H}}\). This completes the proof, yielding the final bound:
\begin{equation}
R(f_\theta,X) \le \widehat{R}(f_\theta, X_s) + M\,\mathfrak{R}(\mathcal{H}) + \sqrt{\frac{\ln(1/\delta)}{2H}}.
\end{equation}
Thus, by defining a modified margin function \(\varphi_{\lambda,h}\) and its smooth surrogate \(\psi_\lambda\), and by incorporating the coreset property into our uniform convergence framework, we have derived a bound on the full-image risk in terms of the empirical risk computed on the coreset, plus a complexity term and a deviation term. It not only ensures that the coreset approximates the full loss well but also tightly controls the additional estimation error.

\section{Benchmark Datasets}
\label{sec_app:dataset}
In this section, we briefly introduce all datasets used in our experiments. In summary, the benchmark datasets can be divided into two categories: (i) scenes image classification, i.e., NWPU-RESISC45 \cite{NWPU-RESISC45}, AID \cite{AID}, and RSI-CB \cite{RSI-CB} datasets; (ii) semantic segmentation, i.e., TikTok dances \cite{roman2023humantiktok}, TrashCan \cite{hong2020trashcan}, and GTEA \cite{fathi2011learning} datasets. The composition of the data set is as follows:
\begin{itemize}
    \item NWPU-RESISC45 \cite{NWPU-RESISC45} contains 31,500 images covering 45 scene categories, with various resolutions for studying multi-scale remote sensing image classification. We select this dataset to verify whether the proposed HCS can enhance discrimination and robustness under different resolutions and scales, especially for wide-area scene classification.
    \item AID \cite{AID} comprises over 10,000 aerial images from different regions, covering 30 scene categories with high diversity. Its cross-regional data collection helps assess whether the generalization ability of VLMs can be activated by HCS across various areas and environments.
    \item RSI-CB \cite{RSI-CB} contains approximately 36,000 scene image patches covering 45 categories. By stitching together patches of similar types, it can simulate wide-area image classification tasks in real-world scenarios.
    \item TikTok dances \cite{roman2023humantiktok} includes 2,615 images of dancers extracted from TikTok videos, providing full-body segmentation annotations. This dataset is used to test the performance of HCS in capturing the collaborative abilities of different regions for tracking human dynamics and pose variations.
    \item TrashCan \cite{hong2020trashcan} collects 1,484 real underwater images annotated with six types of waste. Due to the challenging lighting and visual conditions underwater, we use this dataset to evaluate the model's robustness and segmentation accuracy under extreme environments, thereby improving its performance in complex lighting and background interference.
    \item GTEA \cite{fathi2011learning} comprises seven sets of first-person daily activity videos, totaling approximately 3,500 frames. Featuring occlusions and complex human movements, it is used to assess the ability of HCS to select key regions, handle partial occlusions, and manage complex dynamic scenes, thus reflecting the potential improvements of VLMs.
\end{itemize}

\section{Baselines}
\label{sec_app:baselines}
We briefly introduce the baselines used in the experiments. Note that for the VLMs, we directly use the open-sourced pre-trained weights for implementation and evaluation.
\begin{itemize}
    \item ViT-B/32 \cite{liu2021swin}: A base-size Vision Transformer with a 32×32 patch resolution, widely used as a strong backbone.
    \item ViT-L/14 \cite{liu2021swin}: A larger Vision Transformer with a 14×14 patch resolution that offers more detailed image representations.
    \item CLIP \cite{CLIP}: A model trained on large-scale image–text pairs to learn joint visual and textual representations, useful for zero-shot tasks.
    \item ContextCLIP \cite{grover2022contextclip}: A CLIP variant that incorporates additional contextual information to improve downstream task performance.
    \item LLaVA-hf/llava-v1.6-mistral-7b-hf \cite{liu2023visual}: A vision–language model built on the LLaVA framework with a Mistral 7B backbone, designed for interactive visual conversation.
    \item LLaVA-hf/llama3-llava-next-8b-hf \cite{liu2023visual}: Another LLaVA variant that leverages an 8B-parameter LLaMA-3 model for enhanced visual-language understanding.
    \item Qwen/Qwen2-VL-7B-Instruct \cite{bai2025qwen25vltechnicalreport}: A 7B-parameter instruct-tuned visual–language model from the Qwen series, aimed at robust instruction following in multimodal tasks.
    \item Vanilla SAM \cite{sam}: The original model that enables promptable segmentation across various domains and input types.
    \item Med-SA \cite{wu2023medical}: A specialized version of SAM for medical segmentation that introduces a Space-Depth Transpose to extend 2D segmentation to 3D medical images and uses a Hyper-Prompting Adapter for prompt-conditioned adaptation.
    \item SAMed \cite{tancik2020fourier}: Enhances SAM for medical image segmentation by integrating Fourier analysis to incorporate frequency-domain information, thereby boosting robustness and accuracy.
    \item BLO-SAM \cite{zhangblo}: An improved version of SAM that refines both the segmentation network and task-specific objectives to overcome challenges associated with human annotation.
    \item OVSAM \cite{yuan2024open}: A SAM-based approach focused on open-vocabulary segmentation, leveraging CLIP to augment SAM's capabilities.
    \item CLIPSAM \cite{CLIP}: A SAM variant that exploits CLIP’s image–text representations to provide more semantically meaningful segmentation results.
\end{itemize}
These methods collectively represent a spectrum of approaches to scene understanding, covering two classic tasks, i.e., scene image classification and semantic segmentation.

\section{Implementation Details}
\label{sec_app:implementation}
Our implementation of HCS builds on a simple yet effective three-layer MLP that is fully plug-and-play with any VLM. In our experiments, we refine the interpretable regions of input samples by leveraging the features extracted from the VLMs. Importantly, our approach does not require fine-tuning the VLMs themselves; instead, we train this lightweight network solely for coreset selection, which in turn enhances the overall performance of the baselines. The feature map generated by the baseline VLM is used to determine the salient regions of the input, and the HCS then refines these regions to select a representative coreset. This design enables us to evaluate the VLMs without any additional fine-tuning of their parameters.
For the estimation of feature distribution, such as the calculation in $S_{re}$, we first extract features from $X$ and $X_s$ using a pretrained VLM, represented by the embeddings from the final convolutional layer or the projection head, depending on the VLM architecture. Kernel density estimation (KDE) is then applied to approximate the probability density functions $p(X)$ and $p(X_s)$. The remaining computations are implemented based on the formal expressions mentioned in Subsection \ref{sec:method_function}. For example, for $S_{re}$, the KL divergence of feature distributions is calculated following $D_{KL}(p(X_s)|p(X))=\sum_{i}p(X_s(i))\log\frac{p(X_s(i))}{p(X(i))}$ where $p(X(i))$ and $p(X_s(i))$ denotes the density of sample $i$ under the global/coreset distribution. and the summation is approximated in discrete form after KDE; for $S_{ut}$, we calculate directly based on the loss of the models.
Moving on to the optimization process, we optimize the HCS using the Adam optimizer with a momentum value of 0.8 and a weight decay of \(10^{-4}\). The initial learning rate is set to 0.1 and is linearly scaled when needed to adjust to different experimental conditions. All experiments are run for five independent trials to ensure statistical reliability. The training and evaluation are carried out on 8 NVIDIA Tesla V100 GPUs, ensuring that our results are reproducible and that our method scales well with the computational resources available.

\section{Full Experimental Results and Analyses}
\label{sec_app:experiment}
In this section, we provide the full experimental results and analyses due to space limitations, including more comparisons with other feature-based, coreset-based, and region-based selection methods, trade-off performance, and performance when facing noise.

\subsection{Comparison with Other Selection Methods}
\label{sec_app:experiment_comparison}
To evaluate the performance of HCS, we further construct a comparison with different selection methods.
Specifically, we select two recently proposed SOTA feature and region selection strategies, i.e., sample selection and token selection methods \cite{wang2024data,ataee2023max}, as well as three core-set selection approaches, i.e., standard coreset selection (SCS) \cite{har2004coresets} and newly proposed methods \cite{chai2023efficient,vepa2025integrating}. The evaluation is performed on the NWPU-RESISC45 dataset with ViT-B and CLIP under the same settings. Since we are the first to introduce coreset selection into wide-area scene understanding and these methods were originally developed for other domains, to ensure a fair comparison: (i) For \cite{wang2024data,har2004coresets,chai2023efficient,vepa2025integrating}, we adapt their original sample selection modules to perform region selection and embedded them into the base model. (ii) For \cite{ataee2023max}, we follow its design to constrain the VLM’s attention layers for token-level selection.
The results are shown in \textbf{Table \ref{tab:comparison}}. From the results, we can observe that HCS achieves the highest accuracy with the lowest computational cost for wide-area scene understanding, further demonstrating the advantages of both effectiveness and efficiency.

\begin{table}[t]
\centering
\caption{Comparison with other selection methods.}
\label{tab:comparison}
\begin{tabular}{lcc}
\toprule
\textbf{Method} & \textbf{Accuracy (\%)} & \textbf{Calculation Cost ($N\times$)} \\
\midrule
ViT-B+HCS         & 17.2 & $1\times$ \\
ViT-B+DAUS \cite{wang2024data}    & 13.1 & $3.2\times$ \\
ViT-B+MMTS \cite{ataee2023max}    & 11.9 & $2.9\times$ \\
ViT-B+SCS \cite{har2004coresets}    & 8.6  & $1.7\times$ \\
ViT-B+ESC \cite{chai2023efficient}     & 12.5 & $2.1\times$ \\
ViT-B+al-seg \cite{vepa2025integrating} & 9.2  & $2.4\times$ \\
CLIP+HCS          & 21.4 & $1\times$ \\
CLIP+DAUS \cite{wang2024data}     & 17.3 & $3.8\times$ \\
CLIP+MMTS \cite{ataee2023max}     & 15.2 & $3.4\times$ \\
CLIP+SCS \cite{har2004coresets}     & 9.1  & $1.6\times$ \\
CLIP+ESC \cite{chai2023efficient}      & 15.0 & $1.7\times$ \\
CLIP+al-seg \cite{vepa2025integrating}  & 13.4 & $2.7\times$ \\
\bottomrule
\end{tabular}
\end{table}

\subsection{Model Efficiency and Trade-off Performance}
\label{sec_app:experiment_trade-off}
\begin{figure}
    \centering
    \begin{subfigure}[t]{\linewidth}
        \centering
        \includegraphics[width=\textwidth]{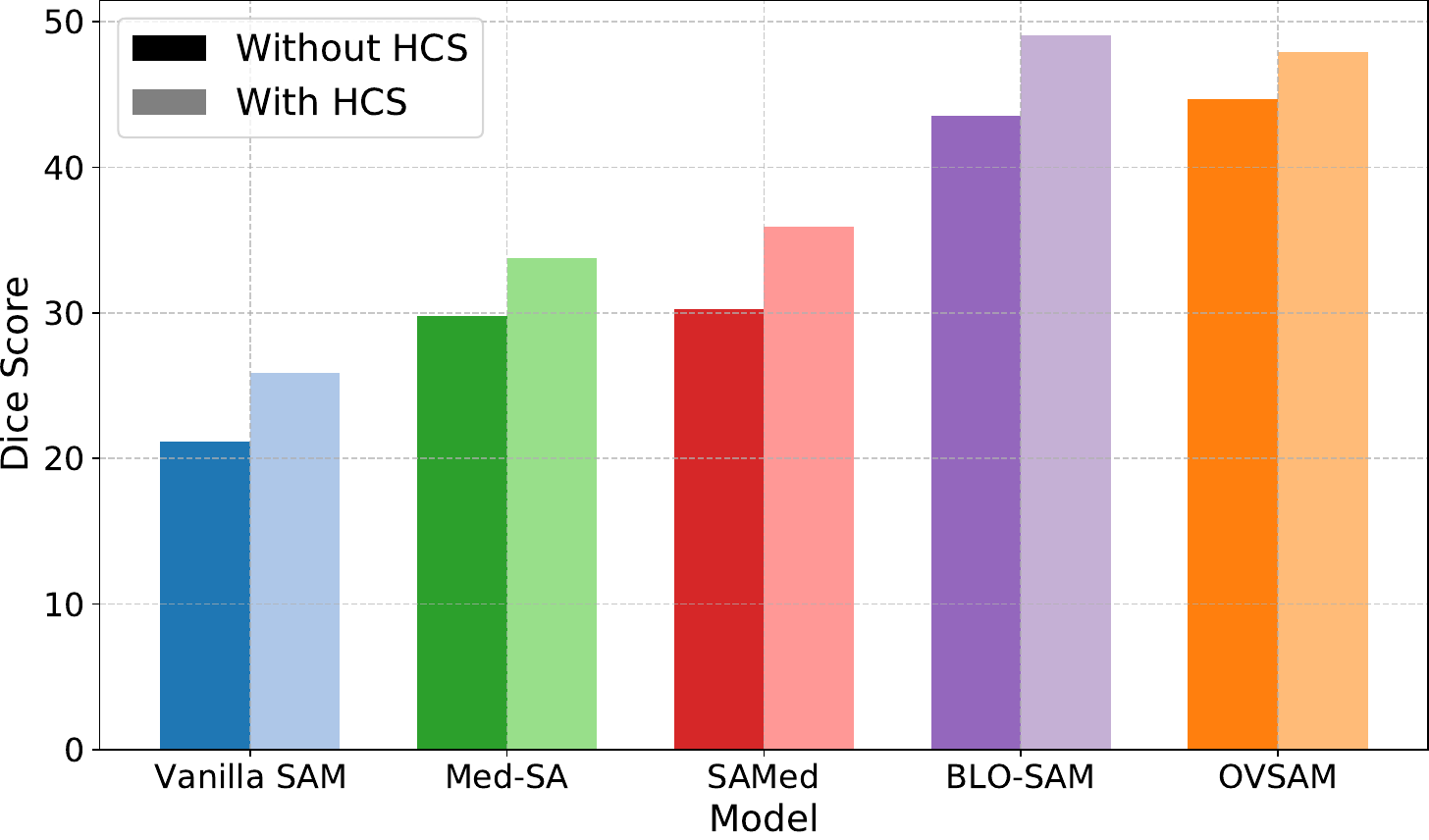}
        \caption{Performance} 
    \end{subfigure}
    \hfill
    \vspace{0.1in}
    \begin{subfigure}[t]{\linewidth}
        \centering
        \includegraphics[width=\textwidth]{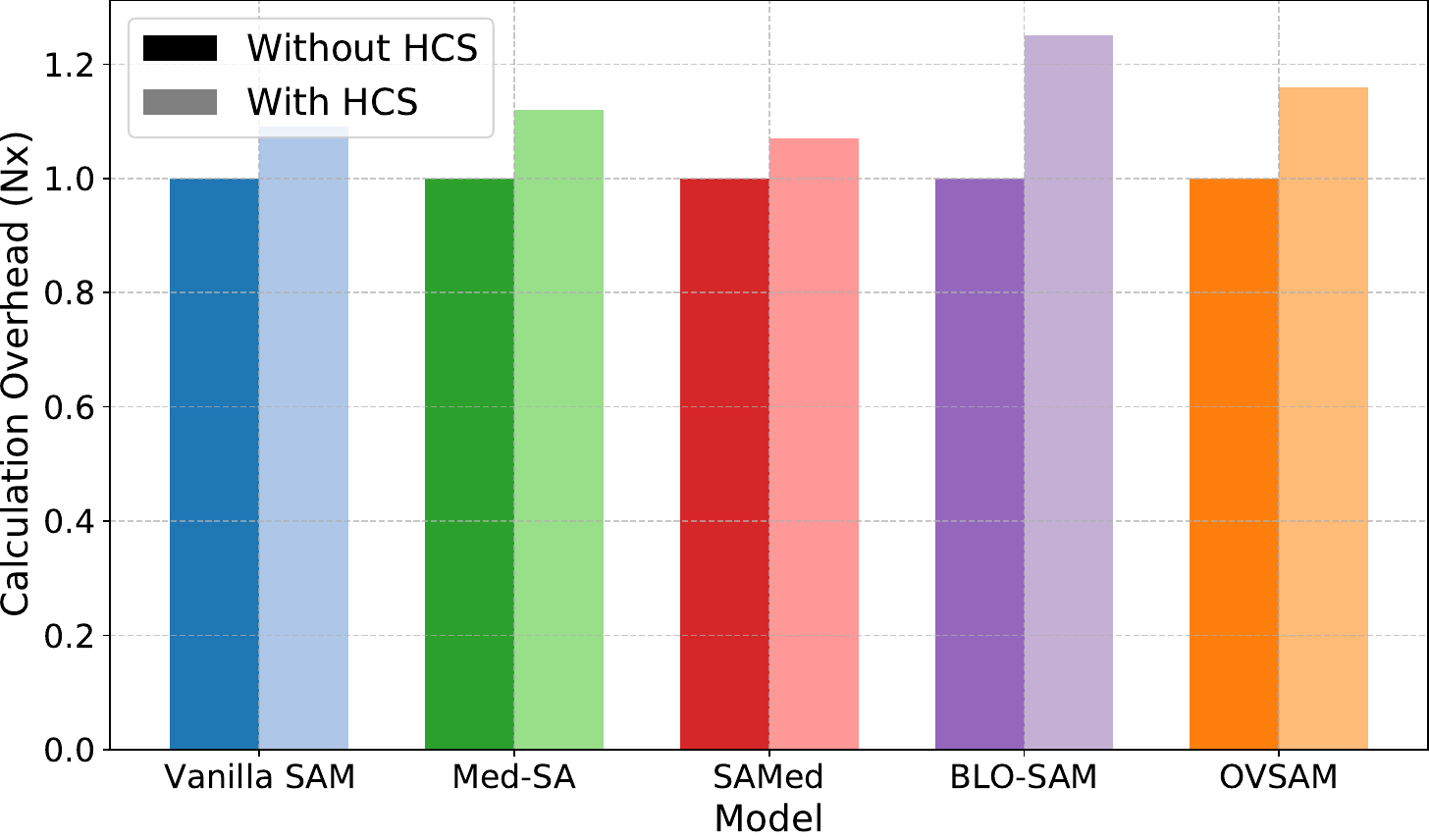}
        \caption{Efficiency} 
    \end{subfigure}
    \caption{Trade-off performance.}
    \label{fig:ex_model_effi}
\end{figure}

Since the proposed HCS is a plug-and-play method, in order to ensure its practicability, we explore the balance between model performance and efficiency. We compare the trade-off performance of multiple baselines before and after using our HCS. We select TrashCan as the benchmark dataset and the semantic segmentation models as the baselines. It is worth noting that since HCS improves the model’s scene understanding performance at test time without requiring fine-tuning, we do not use training time as an evaluation metric. Instead, we assess the per-image inference speed during deployment, using the embedded baseline model as the reference (i.e., $1\times$). The results illustrated in \textbf{Figure \ref{fig:ex_model_effi}} show that introducing HCS achieves great performance with acceptable computational cost (less than 1.3$\times$).

\begin{table}
    \centering
    \caption{Performance comparison (Dice score \%) when facing noises on the TrashCan dataset. Each dataset is split into annotated base and unannotated target classes in a 1:1 ratio, and models are fine-tuned on the base class from the pre-trained weights. The values in parentheses indicate the improvement after introducing HCS.}
    \label{tab:app_noise}
    \resizebox{\linewidth}{!}{
    \begin{tabular}{l|c|c}
        \toprule
        Model & TrashCan & TrashCan with Mask\\
        \midrule
        Vanilla SAM     & 21.15 & 18.21 \\
        Vanilla SAM+HCS & 25.88 (+4.73) & 24.05 (+5.84) \\
        \midrule
        Med-SA        & 29.81 & 26.31 \\
        Med-SA+HCS    & 33.74 (+3.93) & 31.56 (+5.25) \\
        \midrule
        SAMed         & 30.26 & 26.51 \\
        SAMed+HCS     & 35.91 (+5.65) & 32.17 (+5.66) \\
        \midrule
        BLO-SAM       & 43.55 & 40.19 \\
        BLO-SAM+HCS   & 49.04 (+5.49) & 47.00 (+6.81) \\
        \midrule
        OVSAM         & 44.65 & 39.98 \\
        OVSAM+HCS     & 47.91 (+3.26) & 44.89 (+4.91) \\
        \midrule
        CLIPSAM       & 41.22 & 39.74 \\
        CLIPSAM+HCS   & 48.93 (+6.71) & 44.15 (+4.41) \\
        \bottomrule
    \end{tabular}}
\end{table}

\begin{figure}
    \centering
    \includegraphics[width=0.9\linewidth]{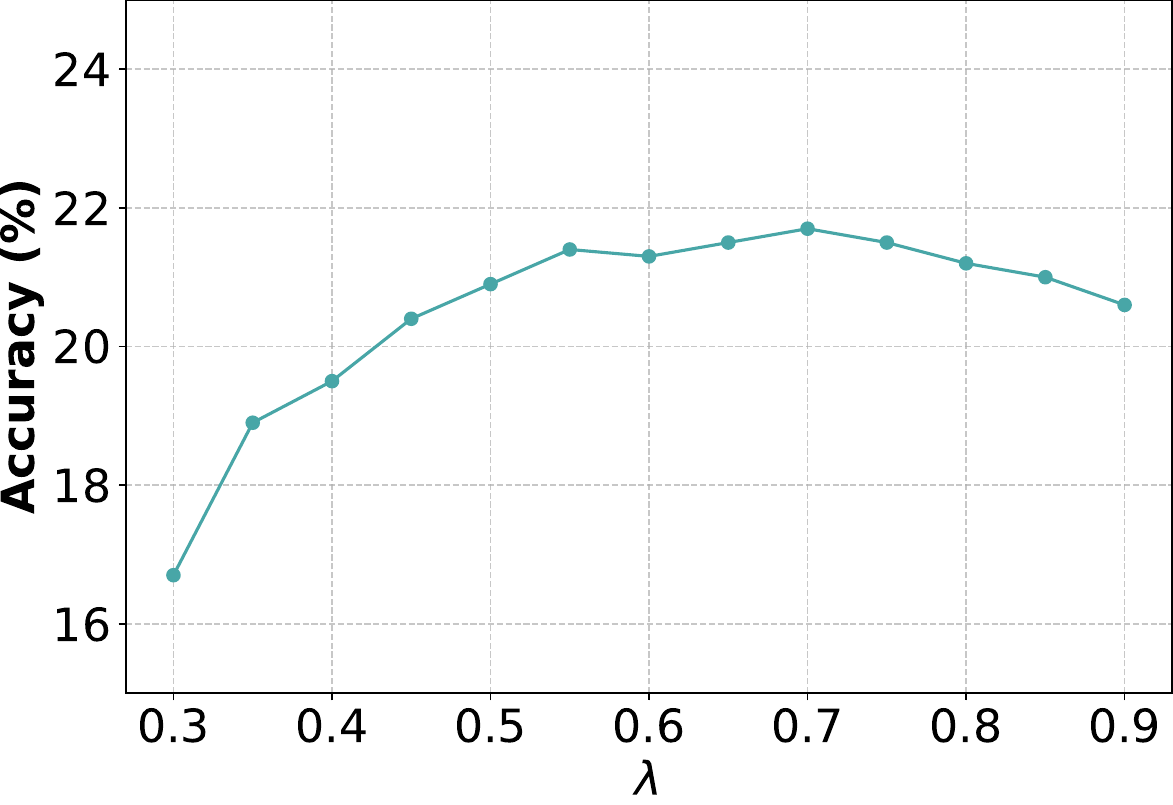}
    \caption{Parameter sensitivity of the hyperparameter $\lambda$ in the score of unility.}
    \label{fig:app_parameter}
\end{figure}

\begin{table}[t]
\centering
\caption{Performance of HCS applied to hierarchical visual backbones and VLMs.}
\label{tab:app_hiera_vlm}
\begin{tabular}{lc}
\toprule
\textbf{Methods} & \textbf{Accuracy (\%)} \\
\midrule
Swin-L       & 2.41 \\
Swin-L+HCS   & 19.95 \\
BLIP-2       & 5.14 \\
BLIP-2+HCS   & 23.58 \\
\bottomrule
\end{tabular}
\end{table}

\begin{table}[t]
\centering
\caption{Effect of hierarchical depth on performance.}
\label{tab:depth}
\begin{tabular}{lc}
\toprule
\textbf{Depth} & \textbf{Accuracy (\%)} \\
\midrule
1         & 8.4  \\
4         & 15.9 \\
$D$       & 21.4 \\
$D_{\text{max}}$ & 20.5 \\
\bottomrule
\end{tabular}
\end{table}

\subsection{When Facing Noise for Scene Understanding}
\label{sec_app:experiment_noise}
Our proposed HCS is designed to help the model better understand images by selecting interpretable regions from wide-area images. The robustness score, \(S_{ro}(X_s)\), incorporates filtering of the effects of noise in the image.
To comprehensively evaluate the effect of HCS in the presence of noise, we further provide a toy experiment to intuitively evaluate the impact of noise on HCS. Specifically, we select TrashCan as the benchmark, since it already contains the image with noises due to the challenging lighting and visual conditions underwater. On the basis of the data, we apply a mask to 30\% of the data area following \cite{wang2024image} to simulate the data corruption scenarios. The results are shown in \textbf{Table \ref{tab:app_noise}}. From the results, we can observe that (i) noises affect the performance of the baselines with the degradation exceeding 4\%; (ii) even in the presence of noise, incorporating HCS into baselines results in a performance drop of less than 2\% while maintaining the great performance gains, further demonstrating the robustness and advantages of HCS.

\subsection{Parameter Sensitivity of $\lambda$ in Utility Score}
\label{sec_app:parameter}
In addition to the parameter sensitivity experiments for $\lambda_{re}$, $\lambda_{ro}$, and $\lambda_{sy}$ presented in the main text, we conducted ablation experiments on the hyperparameter $\lambda$ in the score of utility. We evaluated the effect of varying $\lambda$ over the interval [0.3, 0.9] on model performance. For each hyperparameter $\lambda_{\cdot}$, we initially performed a grid search with a step size of 0.05 to identify a promising range. Subsequently, within that range, we refined the search using a step size of 0.01 and recorded the average results. \textbf{Figure \ref{fig:app_parameter}} displays the ablation results. The findings indicate that the optimal performance is achieved at $\lambda=0.7$, and that values greater than 0.5 have a minimal impact on model performance, demonstrating the ease of tuning in practice.

\subsection{Evaluation on Hierarchical VLMs}
HCS is entirely based on region-level feature representations and loss, without relying on the specific hierarchical structure of the backbone or the length of token sequences. It also requires no modification to existing model architectures, ensuring natural compatibility with various visual backbones and VLMs of different structural designs. We have evaluated HCS on thirteen base models with diverse architectures, covering a wide range from shallow to deep and from simple alignment to explicit multimodal fusion. Results demonstrate that HCS can be seamlessly integrated into all cases and consistently yields significant performance improvements.
To further ensure comprehensive evaluation, we additionally integrate HCS into hierarchical visual backbones and VLMs, such as CLIP+Swin Transformer and BLIP-2. The results in \textbf{Table \ref{tab:app_hiera_vlm}} show consistent performance gains, further validating the effectiveness of HCS.

\subsection{Evaluation of Hierarchical Depths}
As described in Subsection \ref{sec:method_selection}, HCS adopts a top-down hierarchical selection strategy that progressively refines region selection from coarse to fine. At each level, the image is partitioned into $N_p \times N_p$ grids, and the hierarchy depth is dynamically determined based on the resolution of the input wide-area image and the coreset selection criterion. This strategy enables accurate identification of the most critical regions without relying on a predefined depth, while substantially reducing the computational overhead compared to pixel-wise selection (Figures \ref{fig:abla_2}, \ref{fig:vis_heatmap}, and \ref{fig:ex_model_effi}). In other words, HCS can adapt to images of varying resolutions and automatically determine the optimal depth. To evaluate how performance varies with increasing hierarchy depth, we conduct a controlled experiment on the NWPU-RESISC45 dataset, analyzing the impact of different depths on HCS effectiveness. Let $D$ denote the optimal depth dynamically selected by HCS, and $D_{\text{max}}$ the maximum pixel-level depth (i.e., fully refined). We report the performance of HCS at levels 1, 4, $D$, and $D_{\text{max}}$. The results in \textbf{Table \ref{tab:depth}} show that performance improves significantly in the early stages of depth increase but saturates or slightly declines beyond $D$, indicating diminishing returns due to redundant computation.

\end{document}